\documentclass[runningheads]{llncs}
\usepackage[T1]{fontenc}
\usepackage{graphicx}
\usepackage{pifont}
\usepackage{adjustbox}
\usepackage{xcolor}
\usepackage{enumerate}
\usepackage{notoccite}
\usepackage{multirow}
\usepackage{amssymb}
\usepackage{amsmath}
\usepackage{float}
\usepackage{graphicx}
\usepackage{tabularx}

\usepackage{hyperref}

\newcommand{\uinet}{RU-UNet}
\newcommand{\methodName}{SCOPE}

\newcommand{\segmentation}{Pixel-Wise Metrics}
\newcommand{\continuity}{Connectivity Metrics}
\begin{document}
\title{SCOPE: Structural Continuity Preservation for Medical Image Segmentation}
%
%

\newcommand{\authormark}[1]{\textsuperscript{#1}}
\author{Yousef Yeganeh\inst{1,3}\authormark{*}  \and
Azade Farshad\inst{1,3}\authormark{*} \and \\
Goktug Guevercin\inst{1}\authormark{*} \and
Amr Abu-zer\inst{1}\authormark{*} \and
Rui Xiao \inst{1}\authormark{*} \and
Yongjian Tang \inst{1} \and \\
Ehsan Adeli \inst{2} \and
Nassir Navab \inst{1,4}}
\renewcommand{\thefootnote}{\fnsymbol{footnote}}
\footnotetext[1]{Equal contribution}
%
%

\institute{
  \begin{tabular}{cc}
    Technical University of Munich\inst{1} & Stanford University\inst{2} \\ 
    MCML\inst{3} & Johns Hopkins University\inst{4}
  \end{tabular}
}
\maketitle              

\begin{abstract}
Although the preservation of shape continuity and physiological anatomy is a natural assumption in the segmentation of medical images, it is often neglected by deep learning methods that mostly aim for the statistical modeling of input data as pixels rather than interconnected structures. In biological structures, however, organs are not separate entities; for example, in reality, a severed vessel is an indication of an underlying problem, but traditional segmentation models are not designed to strictly enforce the continuity of anatomy, potentially leading to inaccurate medical diagnoses. To address this issue, we propose a graph-based approach that enforces the continuity and connectivity of anatomical topology in medical images. Our method encodes the continuity of shapes as a graph constraint, ensuring that the network's predictions maintain this continuity. We evaluate our method on two public benchmarks on retinal vessel segmentation, showing significant improvements in connectivity metrics compared to traditional methods while getting better or on-par performance on segmentation metrics.

\end{abstract}

\section{Introduction}
Deep learning models for medical image analysis are mostly designed to prioritize the statistical modeling of shapes and textures over the morphology of the medical images. Even though the disruption of organs like vessels is often an indication of an underlying disorder, they are often ignored in favor of overall segmentation performance. The importance of modeling structural integrity becomes more prominent for the diagnosis of sensitive organs, such as the eyes, which contain dense vasculature and pose an even greater challenge. That is why such models are mostly evaluated and applied in analyzing fundus images. Although deep learning-based methods have improved the performance in retinal vessel segmentation \cite{guo2021sa,kamran2021rv,zhou2021study}, they often fail to preserve the continuity of shapes, which is crucial for accurate medical analysis. To address this, we propose a novel graph-based approach that enforces shape continuity for image segmentation.

Previous studies have attempted to incorporate topological information into retinal vessel segmentation. One such approach is to design specific loss functions that penalize the disconnectedness of segmented retinal vessels. For example, Yan et al.~\cite{yan2018joint} proposed joint segment-level and pixel-wise losses, which emphasize the thickness consistency of thin vessels. Shit et al.~\cite{shit2021cldice} developed the Centerline Dice (clDice) for tubular structures, a connectivity-aware similarity metric that improves segmentation results with more accurate connectivity information. Some studies have suggested topological objective functions to improve deep-learning models to produce results that have more similar topology as the ground truth, which help to reduce topological errors \cite{clough2020topologicalBetti,zhang2022topology}.

Other approaches incorporate architectures \cite{zhang2018road,alom2019recurrent,liu2022full} that aim to maintain the spatial information in the image. The shape conservation can be improved by either introducing residual connections \cite{zhuang2018laddernet,li2020iternet} or by using deformation-based architectures that rely on a prior mask with the same topological features as the ground truth. For example, Zhang et al.~\cite{zhang2022topology} proposed the TPSN, in which a deformation map generated from an encoder-decoder architecture transforms the template mask into the region of interest. Wyburd et al.~\cite{wyburd2021teds} also developed TEDS-Net on a continuous diffeomorphic framework.

Structural graph neural networks provide another way to model the connectivity and continuity of biological structures. Shin et al.~\cite{shin2019deep} integrated a graph attention network \cite{velivckovic2017graph} into a convolutional neural network to exploit both local properties and global vessel structures. Li et al.~\cite{9562259} extended such models to segment 3D hepatic vessels and modified the network to learn graphical connectivity from the ground truth directly. Yu et al.~\cite{yu2022vessel} maintained vessel topology by constructing edges and predicting links between different nodes generated from semantic information extracted from U-Net \cite{ronneberger2015u}; however, many medical image segmentation models aim to improve their efficiency in pixel-wise accuracy rather than modeling the connectivity and preserving continuity.
To this end, we propose a novel graph-based approach that enforces shape continuity in medical image segmentation. Our method encodes the continuity of shapes as a graph constraint, ensuring that the network's predictions maintain this continuity. Our proposed graph-based model is agnostic to prior CNN segmentation \cite{simonyan2014very}, as opposed to other graph construction approaches in previous works, such as VGN \cite{shin2019deep}.

\begin{figure}[t]
    \centering
    \includegraphics[width=1.0\textwidth]{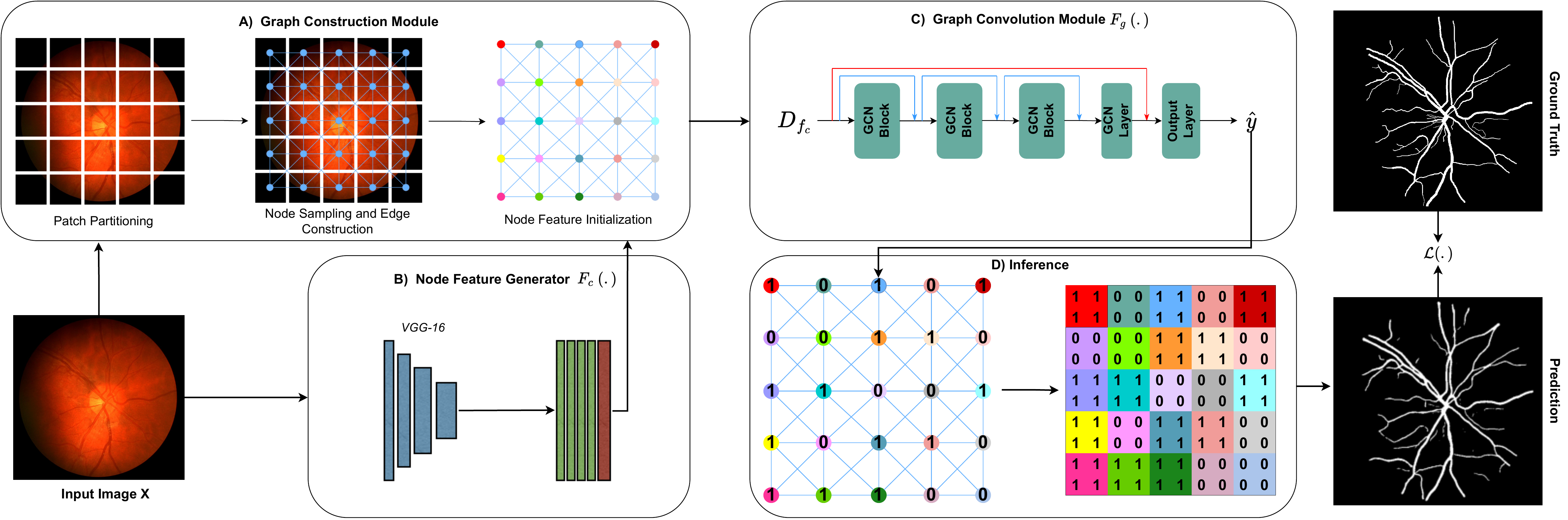}
    \caption{The visual graph is constructed using patches and 1-level edge connections in (A), node features are generated in (B), graph features are extracted in (C) and combined with initial spatial features using skip connections, and final image segmentation is produced in (D) by extending node predictions into corresponding patches.}
    \label{fig:framework}
\end{figure}

Our model directly addresses the issue of preserving shape continuity in medical image segmentation. Unlike previous approaches that rely on selecting the pixel with the highest probability in non-overlapping regions \cite{shin2019deep,velivckovic2017graph,9562259,yu2022vessel}, our method treats the patches in the image as a node for graph construction, resulting in a more efficient approach. We incorporate the DRIU \cite{maninis2016deep} architecture, utilize a graph convolution module during the learning stage, and leverage the clDice loss function \cite{shit2021cldice} to optimize our model performance. Contrary to CNN models, in which the information from the input can affect the inference, by enforcing the relation of neighboring patches of the image via message passing, we inherently bound each part of the input to the neighboring patches affecting the final inference, which our ablation studies on patch size also confirm it. Through comprehensive evaluation, we demonstrate that our approach significantly improves shape continuity metrics compared to the baselines while maintaining comparable or better performance on pixel-wise segmentation metrics. 

%
\section{Methodology}
In this section, we present our continuity preservation network. First, we introduce the overview of the proposed framework (also see \autoref{fig:framework}). Then, we elaborate on the components of our visual graph based on continuity-aware feature generators for the topology of the vessel network.

\subsection{Continuity Preservation Framework}
We design our continuity preservation framework based on graph neural networks (GNNs) \cite{scarselli2008graph} to capture the existing topology of anatomies, such as vessels in the spatial domain. The input image $x \in \mathbb{R}^{H\,\times\,W}$ is divided  into non-overlapping $n \times n$ patches. Each patch, represented as an individual graph vertex $v_i$, connects to its corresponding 1-level neighboring vertices by the edges of a fully-connected graph denoted with the adjacency matrix $A \in \mathbb{R}^{H/n \, \times \, W/n}$, where $H$ and $N$ are the image height and width respectively. The graph representation comprises $N$ vertices, with each vertex responsible for predicting its corresponding $n \times n$ sized patch. To generate initial node-level features for these vertices, inspired by DRIU \cite{maninis2016deep}, we designed a node feature generator $F_c(.)$  that receives the input image, extracts its pixel-wise features, and maps them into node-level features as vertex feature initialization. Then, graph construction module $G(.)$ fuses all sampled graph nodes with their corresponding feature vectors to complete this graph construction process, where $g_{\{A, \, f_{c}\}} = G(x, F_c(x))$. The generated features hold geometrical information about the vertices enabling them to represent general structures of  the organs like vessels; however, they do not enforce any constraints over their relations. To introduce topological information graph convolution module $F_g(.)$ extracts node-level graph features and combines them with their initial spatial features to make final class predictions for the nodes $\hat{y} \in  \mathbb{R}^{N \times 2}$ given ground-truth labels $y \in  \mathbb{R}^{N}$, where $\hat{y} = F_g(A, \, f_{c})$.

\paragraph{Graph Construction}: To construct the visual graph, we first define an image grid of non-overlapping, equally-sized, square $n \times n$ patches, where each patch serves as a vertex of the graph. This results in a vertex set $V = {v_i}^N_{i=1}$, where $N = H/n , \times , W/n$ in an input image $x \in \mathbb{R}^{H ,\times,W}$, as shown in \autoref{fig:framework}-A. Each vertex is descriptive of its corresponding patch and is initialized with feature vectors generated for each of those patches.

We introduce a direct edge construction strategy in which each vertex is connected to all 1-level neighboring vertices of its surrounding patches by undirected and unweighted edges. To capture the neighborhood in every direction, we include not only vertical and horizontal connections but also diagonal edges between the vertices. Previous works \cite{shin2019deep,yu2022vessel} have used geodesic distance to understand the geometry of retinal vessel networks, as it helps identify vertices on vessels that are closer to each other. However, since we sample the vertices everywhere to investigate the entire image region in a granular grid form, we do not use this metric. Instead, we use direct, unconstrained connections, which eliminates the necessity of geodesic vessel interpretation.

\paragraph{Node Feature Generation}: Node feature generation is a crucial step in our architecture, as it maps the pixel-wise features to the corresponding vertex-level features. We follow the approach proposed by Maninis et al. \cite{maninis2016deep}, where a VGG-16 backbone \cite{simonyan2014very} is used to extract the spatial features of a retinal fundus image $x \in \mathbb{R}^{H ,\times,W}$. We obtain a multi-level feature representation $f \in \mathbb{R}^{H ,\times, W ,\times,64}$ from the VGG backbone, capturing both local and global pixel-wise features. To generate the node-level features for our graph, we modify the DRIU architecture \cite{maninis2016deep} by replacing the $1 \times 1$ convolution with a max-pool layer of $n \times n$ window and a stride of $n$, where $n$ is the patch size. This operator downsamples the pixel-wise feature maps to a resolution of $H/n \times W/n$, generating node-level feature maps $f_{c} \in \mathbb{R}^{H/n ,\times, W/n ,\times,64}$. By applying this operator, we obtain a more compact and efficient feature representation for each vertex, which is important for training our segmentation model. \autoref{fig:framework}-B shows the node feature generation process.

\paragraph{Graph Convolution Module} We propose our graph convolution module $F_g(.)$, based on node classification with graph convolution operator \cite{kipf2016semi}, at the top of our grid-based visual graph of the nodes to learn feature representations encoding the relationship and neighborhood between the nodes. First, initial node features $f_c$ are reformatted into node feature matrix $D_{f_c} \in \mathbb{R}^{N \,\times\,64}$ for graph-based processing. Then, $D_{f_c}$ and $A$ are fed into our graph convolution module to generate graph features combined with initial node features by a skip connection, which results in final node feature matrix $D_{f_{g+c}} \in \mathbb{R}^{N \,\times\,96}$. Class prediction for each node, to decide whether the patch of the node contains a vessel branch or not, is made by the last graph convolution layer by using those final node features. There are, in total, eleven graph convolution layers in the network, in which the input to layer $i$ is concatenated with the output of layer $i+2$ by local skip connections to be passed into the layer $i+3$ as shown in \autoref{fig:framework}-C. These skip connections enable the network to have easier gradient flow inside $F_g(.)$ and also between $F_g(.)$ and $F_c(.)$.

\paragraph{Inference} Our inference module (shown in \autoref{fig:framework}-D) is responsible for providing a smooth transition between node predictions and their pixel-wise correspondences. After the graph convolution module classifies each node as foreground or background, the class predictions are extended to $n \times n$ spatial areas that correspond to the nodes in the input image, resulting in a segmentation map in the spatial domain. Our approach differs depending on the size of the node patches. If we use $1 \times 1$ patches, each pixel is treated as a node, and the node predictions are directly accepted as pixel predictions for the segmentation map. However, if we use larger patch sizes such as $2 \times 2$, each node represents a region of pixels. In this case, we use nearest neighbor interpolation to map the class of each node to its corresponding $2 \times 2$ region. This allows us to produce a pixel-wise segmentation map even when the node patches are larger than a single pixel. This way, our mapping function serves as a transition from the graph domain to the spatial domain, which facilitates the interpretation of visual graphs in the image space.


\subsection{Losses} clDice \cite{shit2021cldice} has proven to be an effective loss for preserving continuity; therefore, we employ it for training our graph-based approach. The general idea lying behind clDice loss is to measure the centerline connectivity on the spatial domain from the intersection of segmented and ground-truth vessels with their skeletons. This makes graph-based approaches inconvenient to be trained with clDice. To alleviate this problem, we introduce a reshaping function that puts node predictions of regular visual graphs into a 2D image form to imitate the shape and structure of vessels.


\section{Experiments and Results}
This section provides a comprehensive evaluation of the proposed method for segmenting retinal vessels based on various metrics. The evaluation is carried out on the CHASE\textunderscore DB1 \cite{fraz2012ensemble} and FIVES~\cite{jin2022fives} datasets. We compare our methodology with existing studies, perform an ablation study of graph construction components, and use baseline models such as DRIU \cite{maninis2016deep} and a more recent adaption of UNet (\uinet{}) \cite{kerfoot2019left}. We also evaluate the continuity of vessel segmentations using pre-trained weights from previous studies, including SGL \cite{zhou2021study} and FR-UNet \cite{liu2022full} for the CHASE dataset \cite{fraz2012ensemble}.

\begin{table}[htbp]
    \centering
    \caption{Qualitative results on FIVES \cite{jin2022fives} dataset}
    \label{figs:qualitative} \vspace{-10pt}
    \resizebox{\textwidth}{!}{
    \begin{tabular}{c@{\hskip 0.10cm} c@{\hskip 0.10cm} c@{\hskip 0.10cm} c@{\hskip 0.10cm} c@{\hskip 0.10cm}}
        \hline
        Segmentation Map & GT & \methodName{} (Ours) & DRIU \cite{maninis2016deep} & \uinet{} \cite{ronneberger2015u} \\
        \hline
        \multirow[b]{2}{*}[5.2em]{
        \includegraphics[width=4.3cm,height=4.3cm]{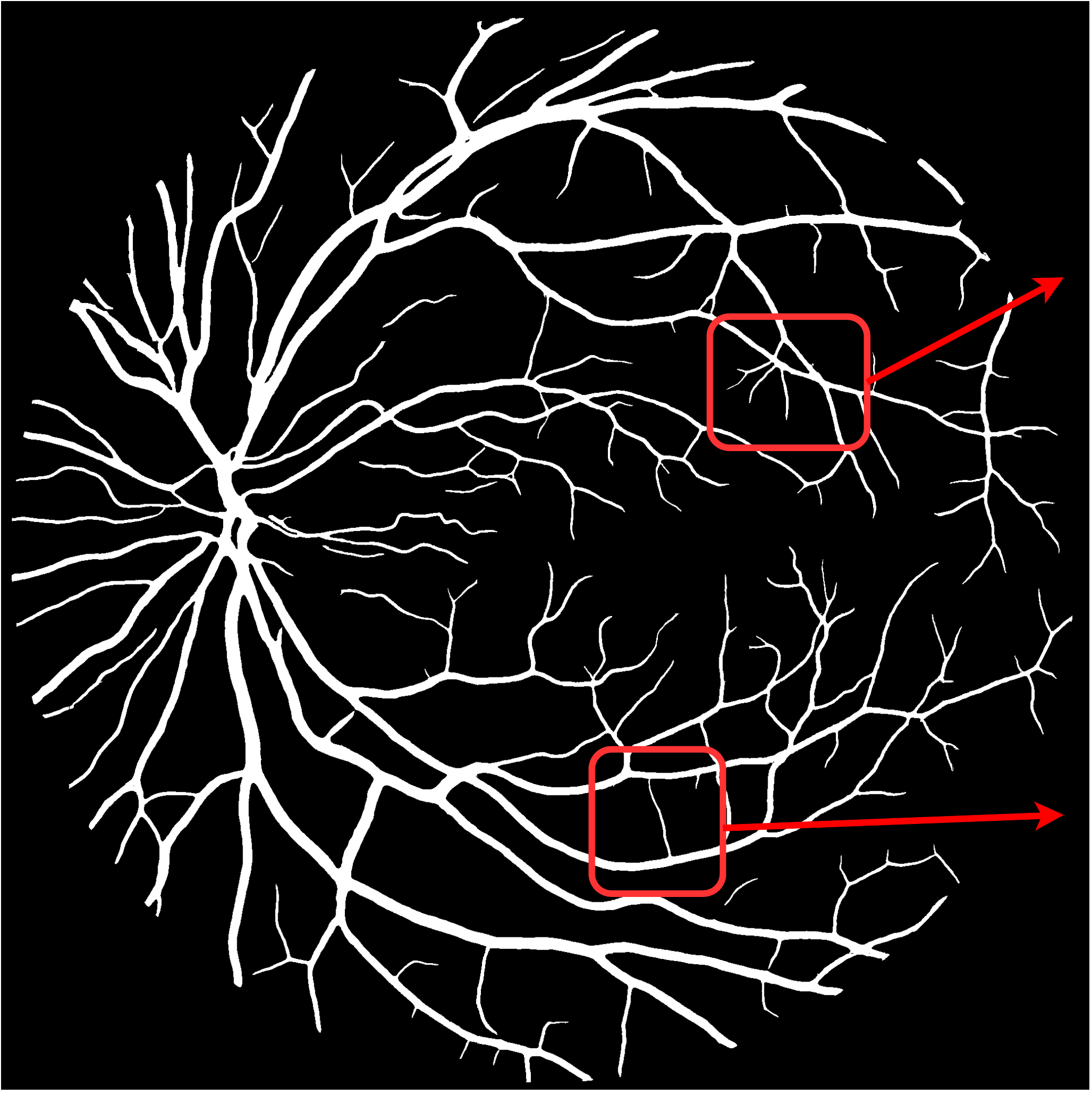}}
        & \includegraphics[width=2.1cm]{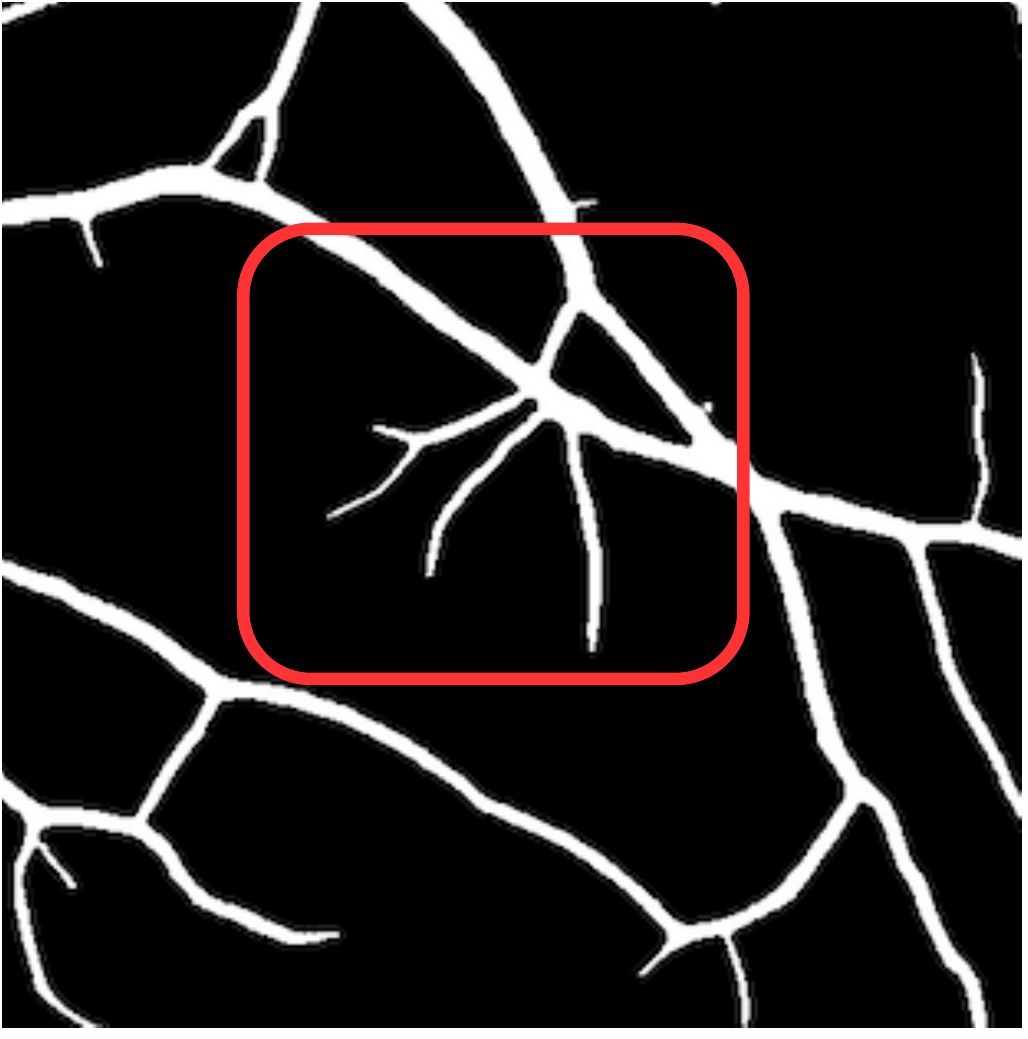} & \includegraphics[width=2.1cm]{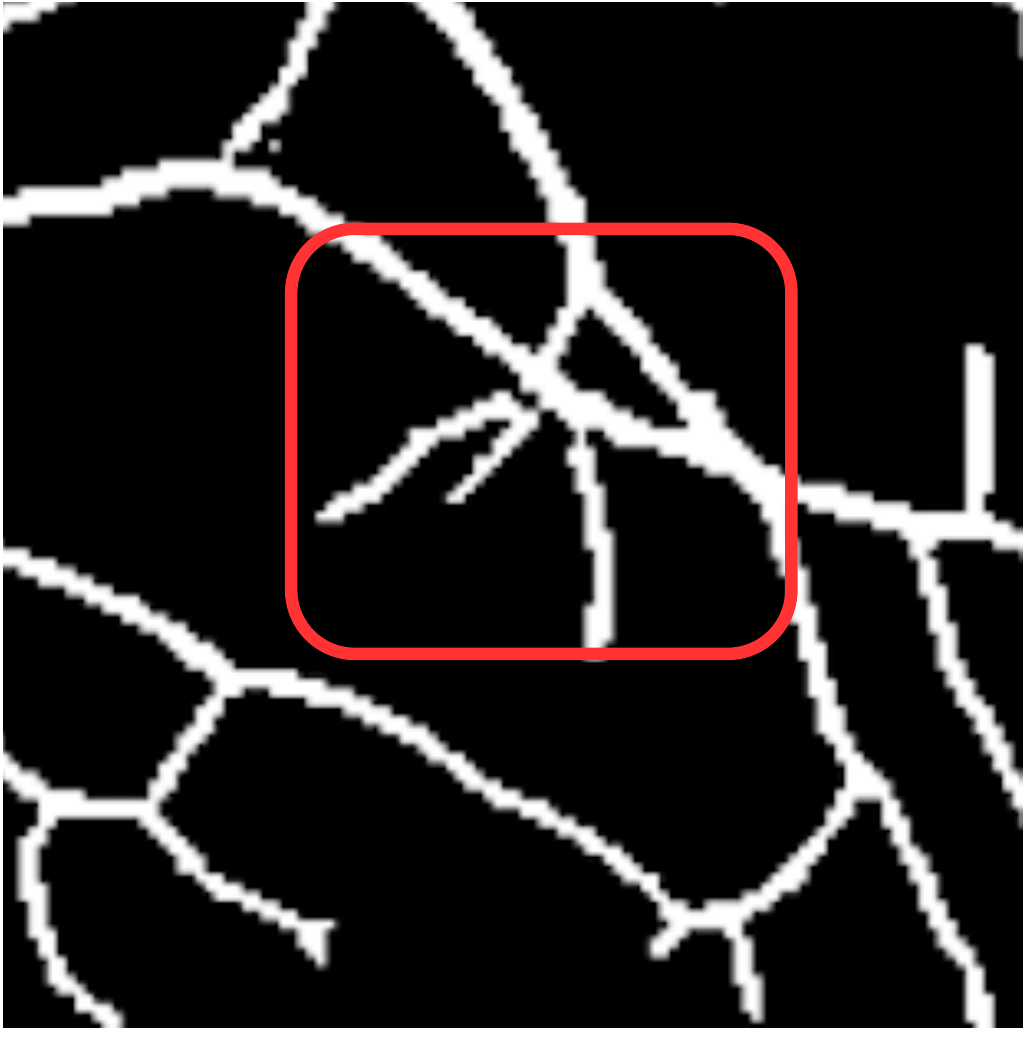} & \includegraphics[width=2.1cm]{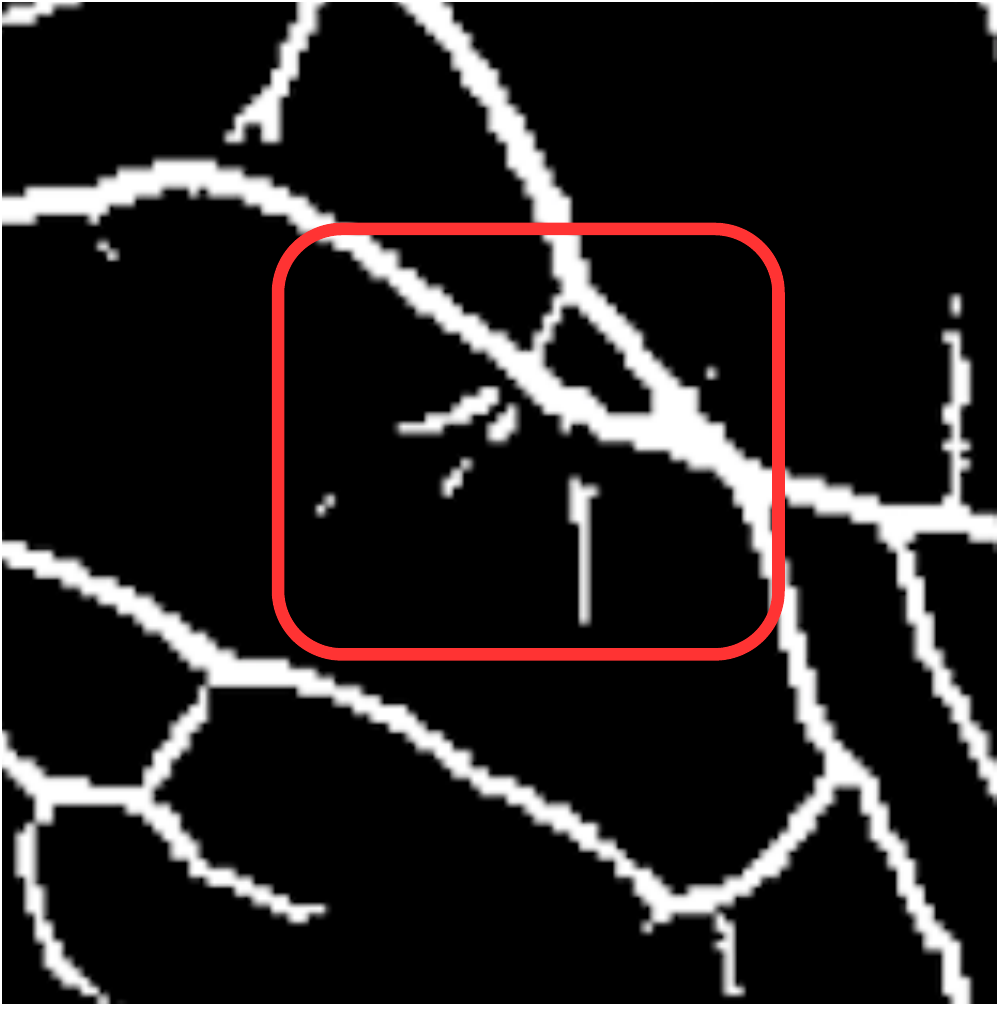} & \includegraphics[width=2.1cm]{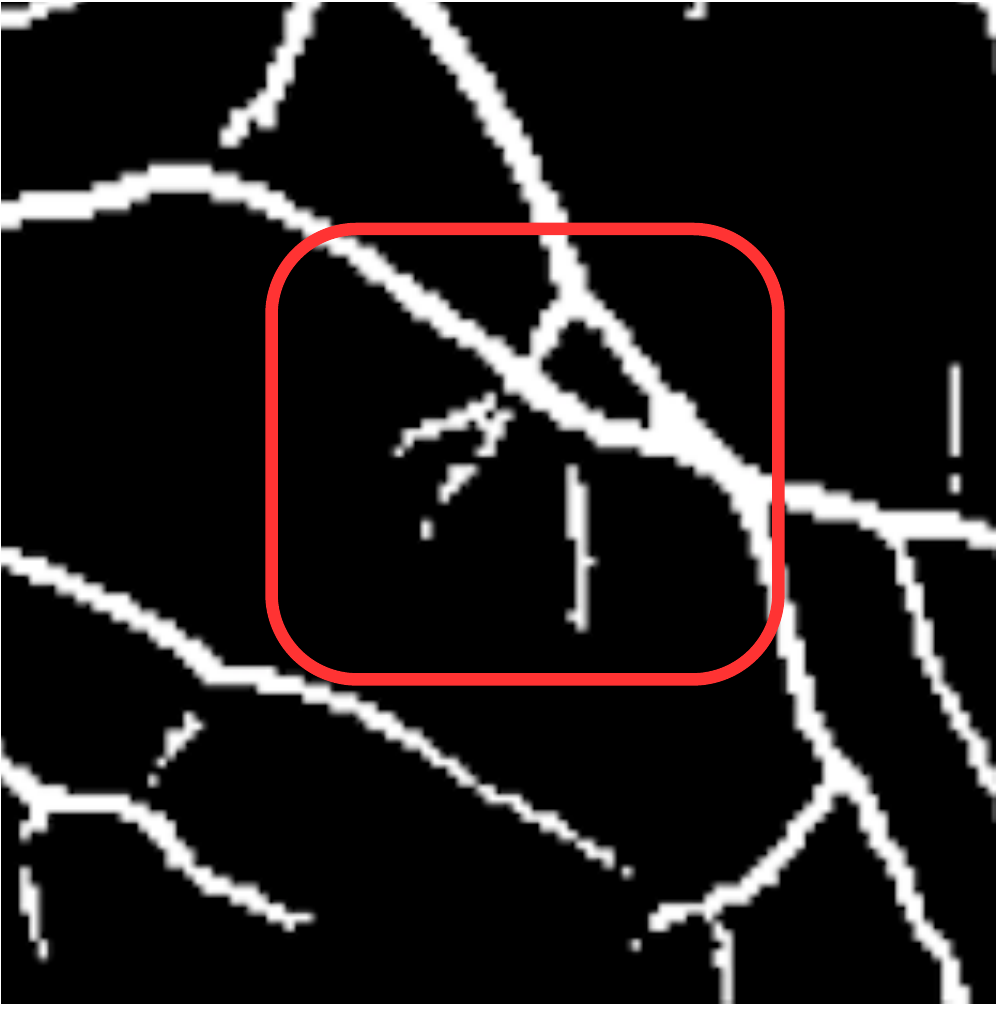} \\
        & \includegraphics[width=2.1cm]{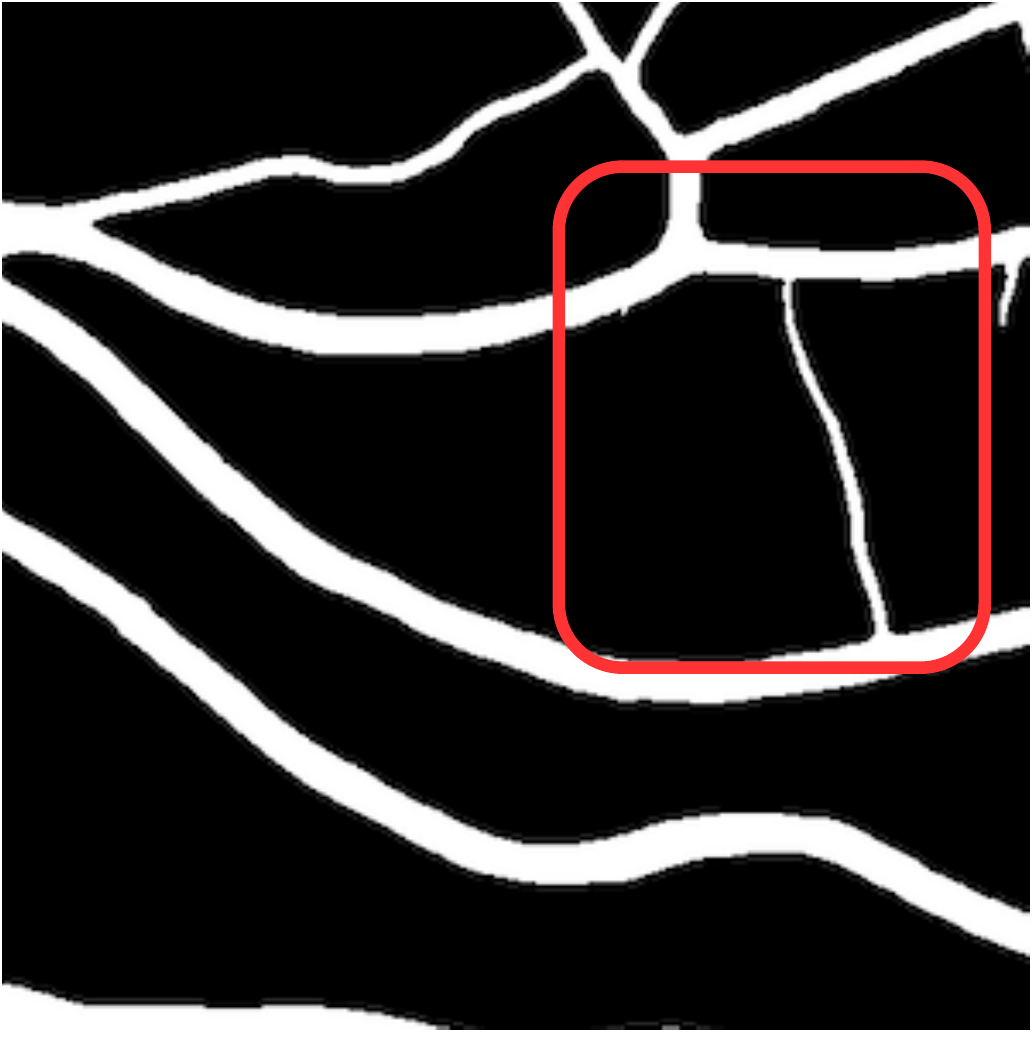} & \includegraphics[width=2.1cm]{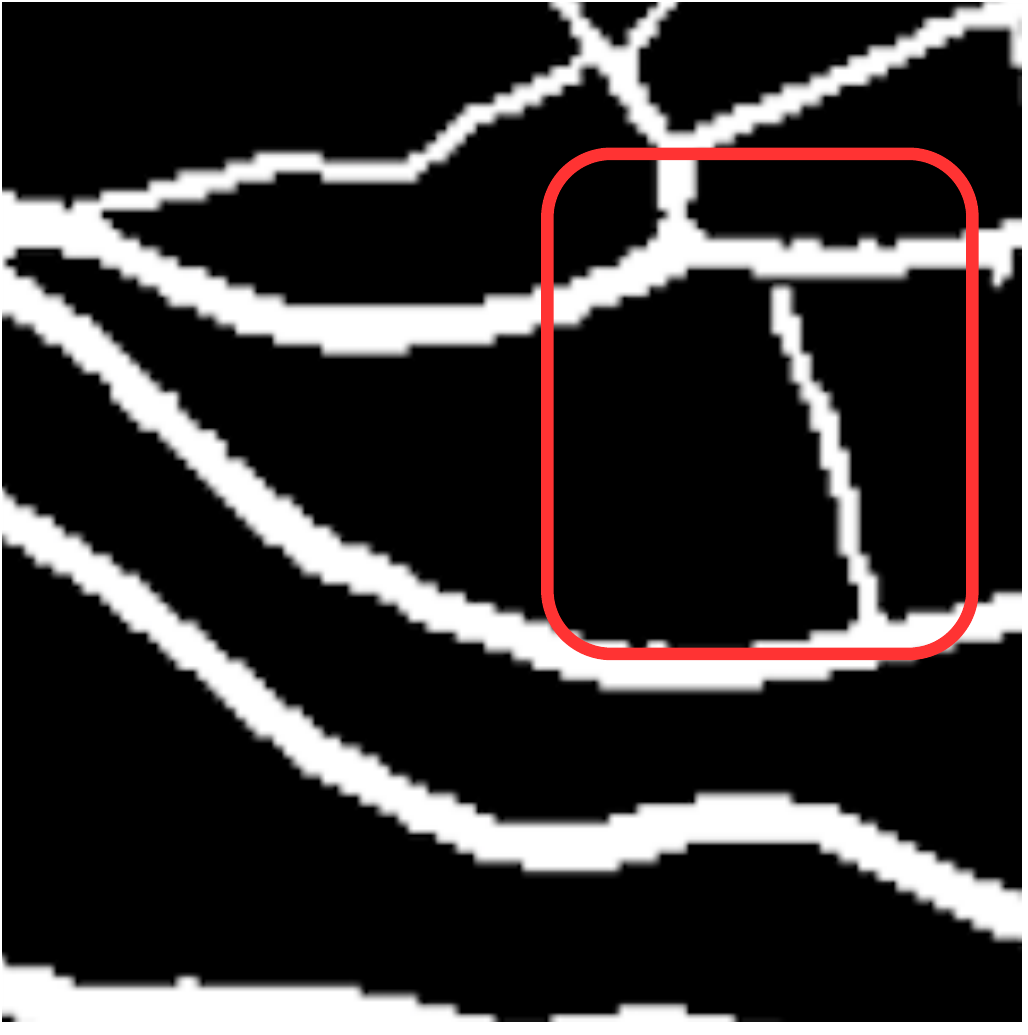} & \includegraphics[width=2.1cm]{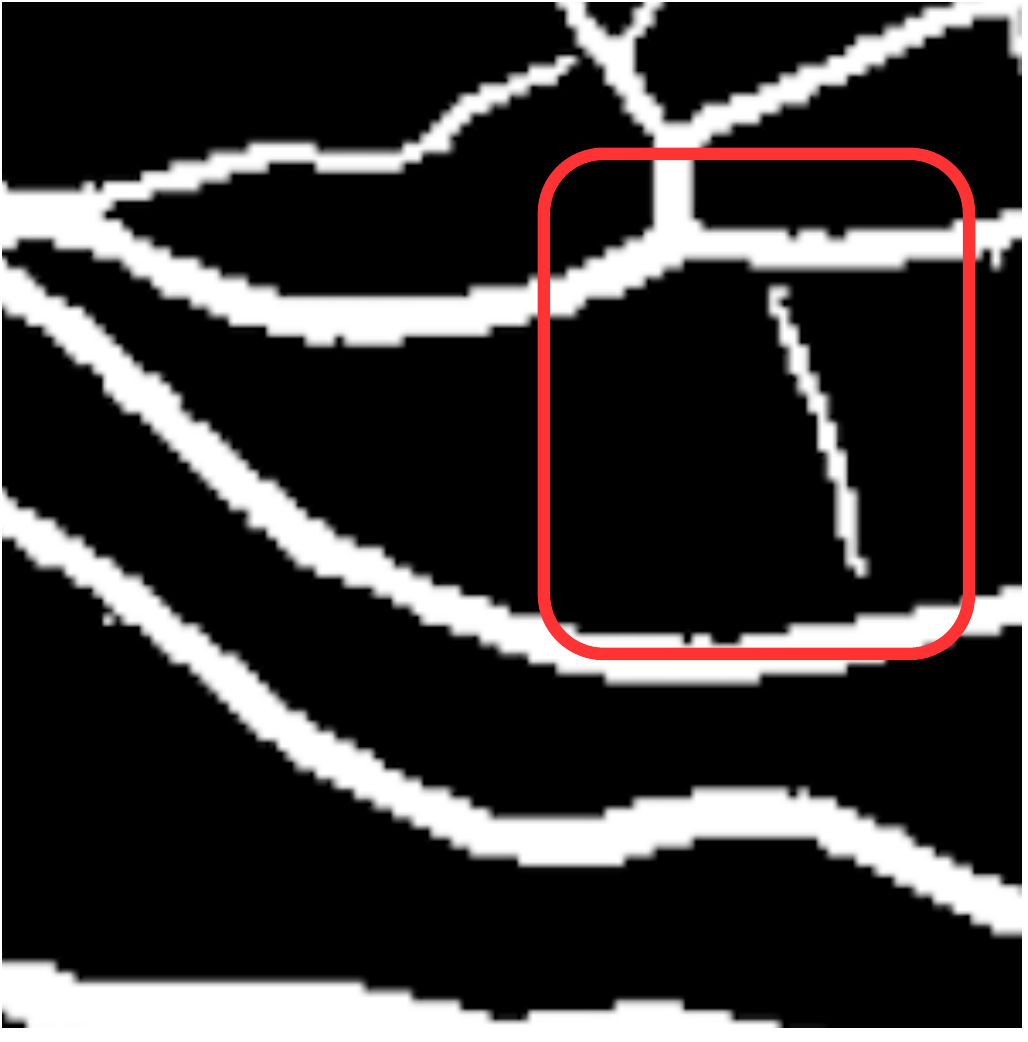} & \includegraphics[width=2.1cm]{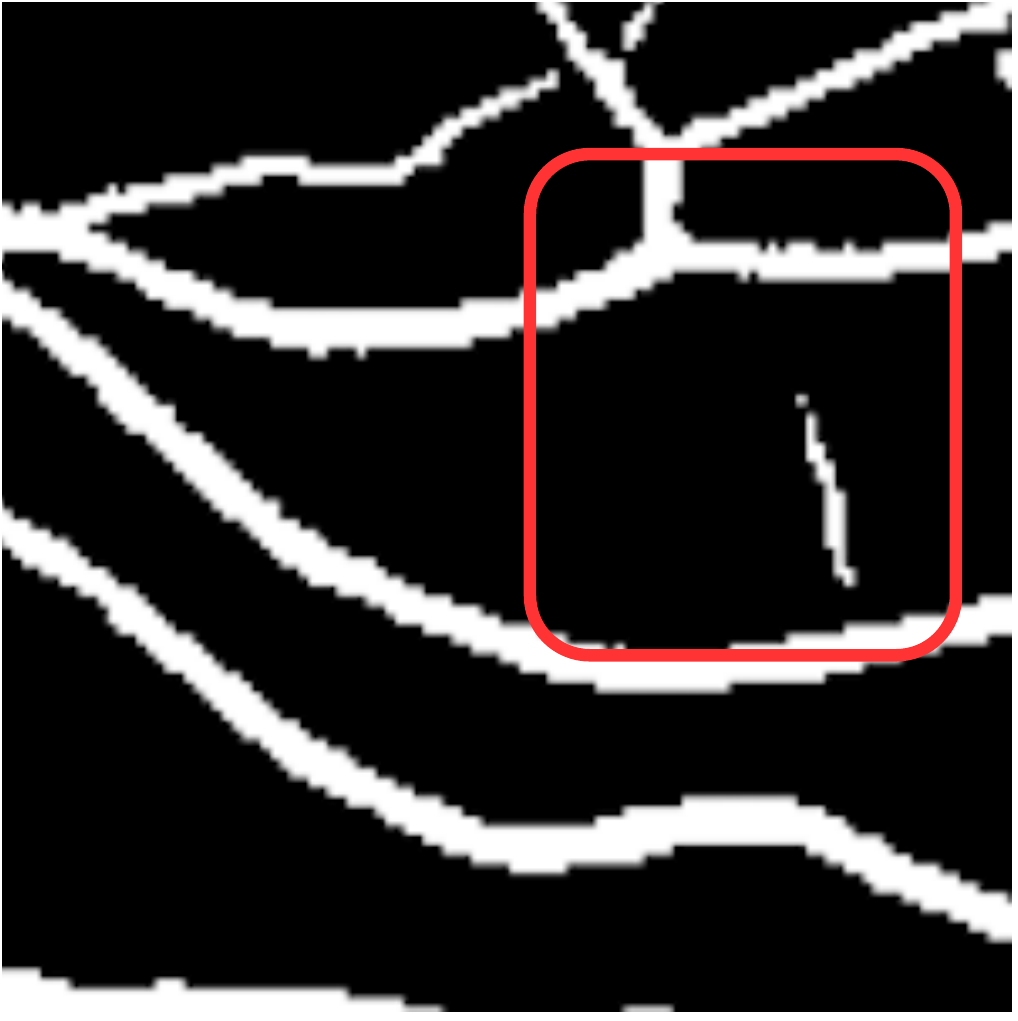}
    \end{tabular}
    }
\end{table}

\begin{table}[tbh]
\caption{Comparison to SOTA on CHASE \textunderscore DB1~\cite{fraz2012ensemble}.}
\label{tab:chase_comparison} \vspace{-10pt}
\centering
\resizebox{\textwidth}{!}{
\begin{tabular}{|l|c c c |c c c c|}
\hline
 &  \multicolumn{3}{c|}{\segmentation{}} & \multicolumn{4}{c|}{\continuity{}} \\ \hline
Method     & Pre (\%) $\uparrow$& Rec (\%) $\uparrow$ &Dice(\%)$\uparrow$& clDice (\%) $\uparrow$ & $error_{\beta_{0}}$ $\downarrow$ & $error_{\beta_{1}}$ $\downarrow$ & $error_{\chi}$ $\downarrow$\\ \hline
\uinet{} \cite{kerfoot2019left}         &     0.78     &  0.80    & 0.79   &    0.74         &            59.8 & 2.8     &    61.2           \\ 
DRIU \cite{maninis2016deep}          & 0.79         &0.80    &  0.79    &0.74           &37.8& 4.2                 &35.1               \\ 
FR-UNet \cite{liu2022full} & 0.76 & \textbf{0.88} & 0.80 & 0.73 & 61.0 & 2.8 & 64.4 \\ 
SGL \cite{zhou2021study} & \textbf{0.79} & 0.87 & \textbf{0.82} & 0.75 & 42.6 & 2.3 & 46.0 \\ 
\methodName{} (Ours) &    0.75      &      0.86  & 0.80 &      \textbf{0.81}       &    \textbf{24.2}& \textbf{1.6}             &     \textbf{22.7}          \\ \hline
\end{tabular}
}
\end{table}

\subsection{Datasets}
\textbf{CHASE\textunderscore DB1}\cite{fraz2012ensemble} has 28 fundus images of size 999 x 960. We split the dataset into two subsets, with the first 20 used for training and the last 8 for testing.

\noindent\textbf{FIVES} \cite{jin2022fives} is composed of 800 fundus images with 2048$\times$2048 resolution split into four equally-distributed subcategories of different diseases: Age-related Macular Degeneration (AMD), Diabetic Retinopathy (DR), Glaucoma, and Normal (see \autoref{figs:sampleFives}). Train and test sets consist of 600 and 200 images, respectively. 



\subsection{Experimental Setup}
All values reported for the \uinet{} \cite{kerfoot2019left} and DRIU~\cite{maninis2016deep} experiments are based on our own implementation. We train the baseline models as well as ours on the FIVES~\cite{jin2022fives} dataset for 100 epochs, using Adam optimizer with a batch size of 8 and weight decay of $1e^{-3}$. We set the learning rate to $1e^{-2}$ for \uinet{} \cite{alom2019recurrent} and $1e^{-3}$ for DRIU~\cite{maninis2016deep}. To address the domain shift problem and adapt the pre-trained weights to the context of CHASE~\cite{fraz2012ensemble}, we fine-tuned the baseline models that are pre-trained on FIVES \cite{jin2022fives}. We used the same hyperparameters for both datasets unless specified. The models were trained on CHASE for 200 epochs. To be consistent with the original paper, we used cross-entropy loss for both datasets during training. We downsample the images to $512 \times 512$ resolution for training and testing in both datasets.

For our proposed method, the node feature generator is trained individually on each dataset using the same hyper-parameters as the DRIU \cite{maninis2016deep} baseline. We incorporate the node feature generator into our graph convolution module, and it is jointly trained with graph-integrated clDice loss \cite{shit2021cldice}. We set the learning rate for training \methodName{} on FIVES to $4e^{-4}$, and the weight decay to $5e^{-3}$. For CHASE \cite{fraz2012ensemble}, we adhere to the same configuration as in FIVES \cite{jin2022fives} except for the number of training epochs which is 200 due to the limited size of the dataset. 
\paragraph{Evaluation Metrics}
Given that the intended goal in this paper is to evaluate the continuity and topology of vessels rather than pure pixel-wise metrics, we use four main topological metrics in our results: (i) clDice \cite{shit2021cldice} evaluating the topological continuity of tubular structures, (ii) $\beta_0$ (Betti zero) \cite{clough2020topologicalBetti} that counts the number of connected components in a topological space,  (iii) $\beta_1$ (Betti one) \cite{clough2020topologicalBetti} indicating the number of independent loops or cycles in the space, and   (iv) $\chi$ (Euler) characteristic \cite{beltramo2021euler} that is a topological invariant quantifying the connectivity and complexity anatomical structure by counting the number of its connected components, holes, and voids, we utilize them as $error_{\beta_{0}}$, $error_{\beta_{1}}$, and $error_{\chi}$ to compare ground truth with predicted results. 

\subsection{Results}
Comparison of vessel segmentation models on CHASE dataset \cite{fraz2012ensemble} using pixel-wise and connectivity metrics shown on \autoref{tab:chase_comparison}, indicates that our model significantly outperforms other methods in continuity metrics and comparable pixel-wise metrics values, outperforming state-of-the-art architectures and two baseline models, a recent modification of \uinet{} \cite{alom2019recurrent}, and DRIU~\cite{maninis2016deep}.


\begin{table}[tb]
\centering
\caption{\textbf{Comparison with previous work.} We present the results of our methodology compared to previous work on the FIVES \cite{jin2022fives} dataset.}
\label{tabs:fives} 
\resizebox{\textwidth}{!}{
\begin{tabular}{|l|l|c c c|c c c c|}
\hline
& & \multicolumn{3}{c|}{\segmentation{}} & \multicolumn{4}{c|}{\continuity{}} \\ \hline
Class           & Architecture &Pre (\%) $\uparrow$& Rec (\%) $\uparrow$ & Dice(\%) $\uparrow$ & clDice (\%) $\uparrow$ & $error_{\beta_{0}}$ $\downarrow$ & $error_{\beta_{1}}$ $\downarrow$ & $error_{\chi}$ $\downarrow$\\ \hline
\multirow{3}{*}{AMD} &\uinet{} \cite{kerfoot2019left}&    0.88        &           0.93&  0.90  &    0.87&        72.8&    4.3&    71.1   \\  
                   &DRIU \cite{maninis2016deep}&              0.87&           \textbf{0.95}&  \textbf{0.91} &  0.87   &        57.8&    2.7&    57.1   \\ 
                   & \methodName{} (Ours)&              \textbf{0.90}&           0.92&  \textbf{0.91}   &   \textbf{ 0.94}&        \textbf{6.7}&   \textbf{2.6}&   \textbf{ 7.9}   \\ \hline
\multirow{3}{*}{DR} &\uinet{} \cite{kerfoot2019left}&              0.85&           0.92& 0.89  &     0.84&        89.5&    4.1&    83.8   \\ 
                   &DRIU \cite{maninis2016deep}&              0.85&           \textbf{0.94}&  0.89  &    0.85&      64.3&    3.8 &60.5  \\  
                   &\methodName{} (Ours)&              \textbf{0.91}&           0.89&  \textbf{0.90}   &   \textbf{0.92}&        \textbf{10.6}&    \textbf{3.0}&    \textbf{14.0 }  \\ \hline
\multirow{3}{*}{Glaucoma} &\uinet{} \cite{kerfoot2019left}             &           0.84&        0.90&   0.87   &  0.83&    87.6&    5.6&83.2   \\ 
                   &DRIU \cite{maninis2016deep}             &           0.87&        \textbf{0.92}&  \textbf{0.89}   &     \textbf{ 0.85}&    51.2&    \textbf{3.4}&   53.1\\ 
                  &\methodName{} (Ours)              &          \textbf{0.91}&        0.77&   0.81   &   0.83&    \textbf{5.4}&  4.5&   \textbf{11.9}\\ \hline
\multirow{3}{*}{Normal} &\uinet{} \cite{kerfoot2019left}            &         0.87&        0.90&    \textbf{0.88}  &    0.84&    141.6&    8.8&123.3   \\ 
                   &DRIU \cite{maninis2016deep}             &          0.83&       \textbf{ 0.94}&    \textbf{0.88}   &   0.84&    138.9&    6.4&   124.8\\ 
                   &\methodName{} (Ours)              &         \textbf{0.89}&        0.87&  \textbf{0.88}   &   \textbf{0.89}&    \textbf{14.0}&    \textbf{4.2}&   \textbf{13.2}\\
                   \hline
\multirow{3}{*}{Average} &\uinet{} \cite{kerfoot2019left}             &         0.85&        0.90&    0.87  &    0.83&    98.0&    5.6&90.5   \\  
                   &DRIU \cite{maninis2016deep}              &          0.85&       \textbf{ 0.92}&    \textbf{0.88}   &   0.84&    73.8&    4.5&   70.4\\ 
                   &\methodName{} (Ours)              &         \textbf{0.90}&        0.85&  0.85   &   \textbf{0.87}&    \textbf{12.1}&    \textbf{3.7}&   \textbf{14.4}\\ \hline
\end{tabular}
}
\end{table}

We also validate our proposed method on the newly released FIVES dataset \cite{jin2022fives} in \autoref{tabs:fives}. To the best of our knowledge, no method has reported results on the FIVES \cite{jin2022fives} dataset. To this end, we compare our model against two competitive models based on the results obtained on CHASE~\cite{fraz2012ensemble}, namely \uinet{} \cite{kerfoot2019left}, and DRIU \cite{maninis2016deep}, as the baselines on FIVES \cite{jin2022fives}. 
The results in \autoref{tabs:fives} show that our model significantly outperforms the two baselines in connectivity metrics $error_{\beta_0}$, $error_{\beta_1}$, $error_{\chi}$, and clDice, while achieving better or comparable results in pixel-wise metrics. This improvement in connectivity metrics indicates that our model enforces the preservation of topological information, while other methods do not strictly enforce it. Our qualitative results in \autoref{figs:qualitative} also support that where other baselines introduced discontinuity to the topology, our method preserves it. Note that since pixel-wise metrics are only sensitive to the pixels, they cannot represent connectivity.

\begin{table}[t]
    \centering
    \caption{Sample Images and Ground-Truth in Fives Dataset}
    \label{figs:sampleFives}\vspace{-10pt}
    \begin{tabular}{c c c c c c}
        \hline
        & AMD & DR & Glaucoma & Normal \\
        \rotatebox{90}{Image} & \includegraphics[width=2cm]{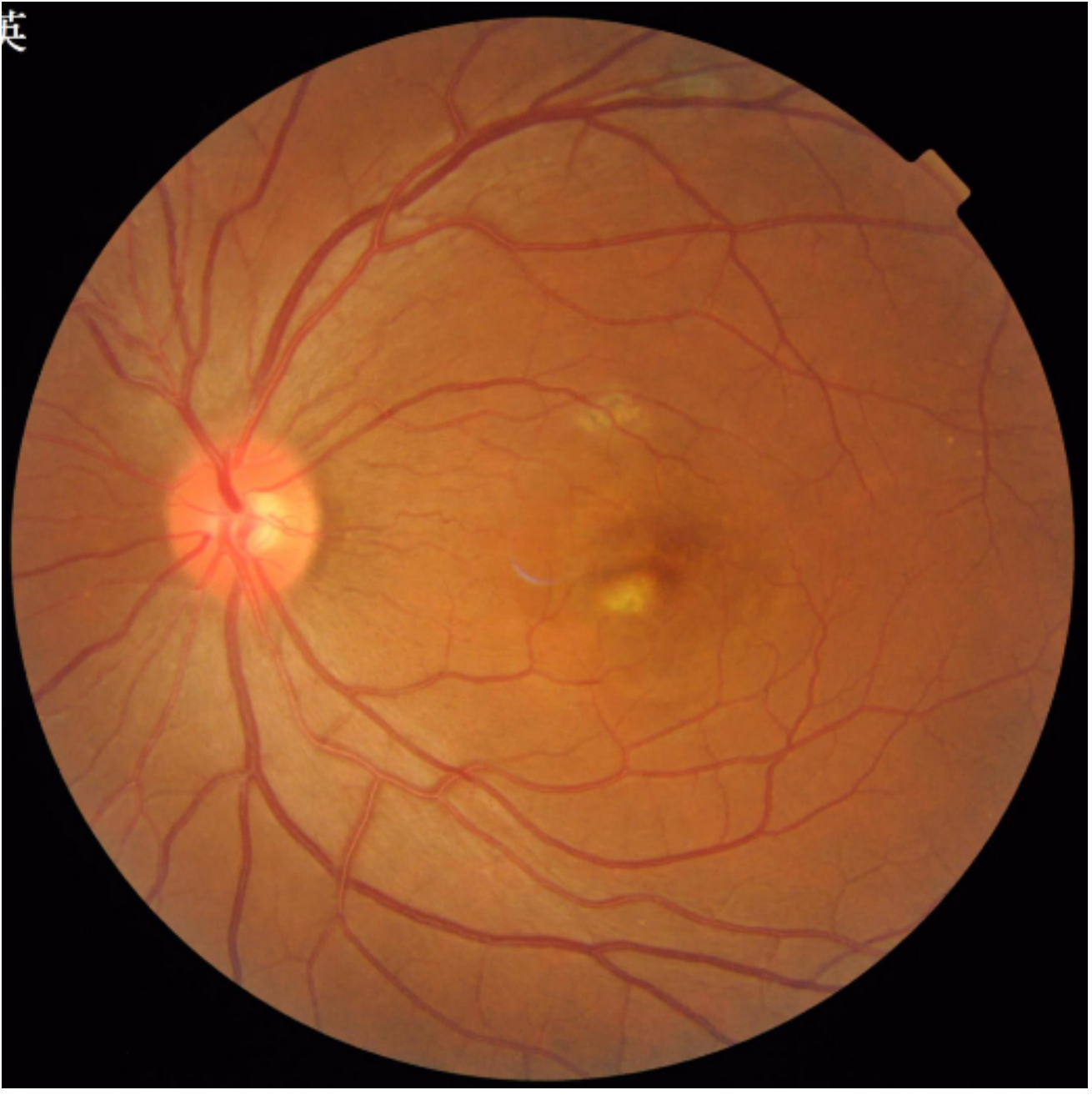} & \includegraphics[width=2cm]{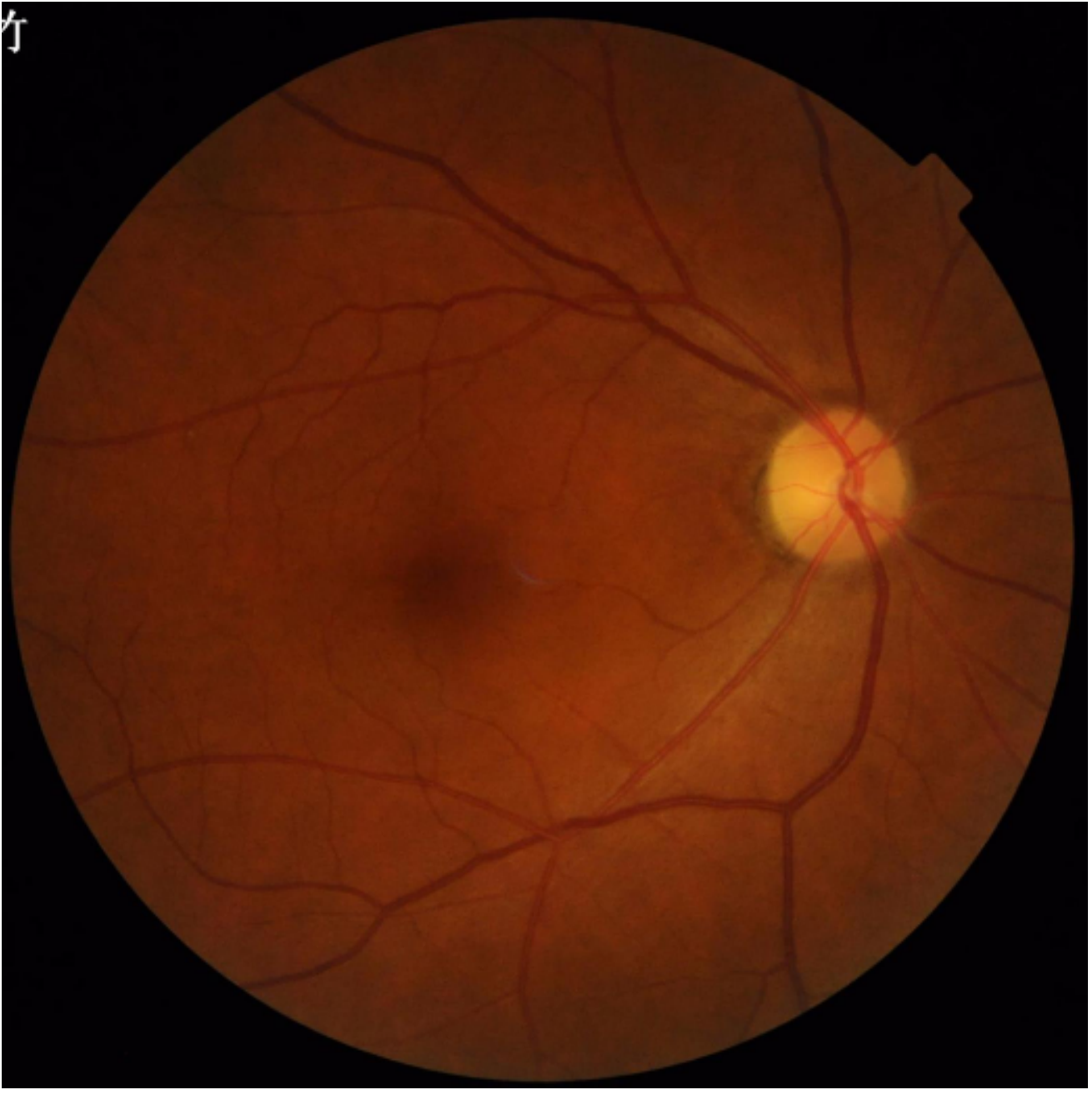} &
        \includegraphics[width=2cm]{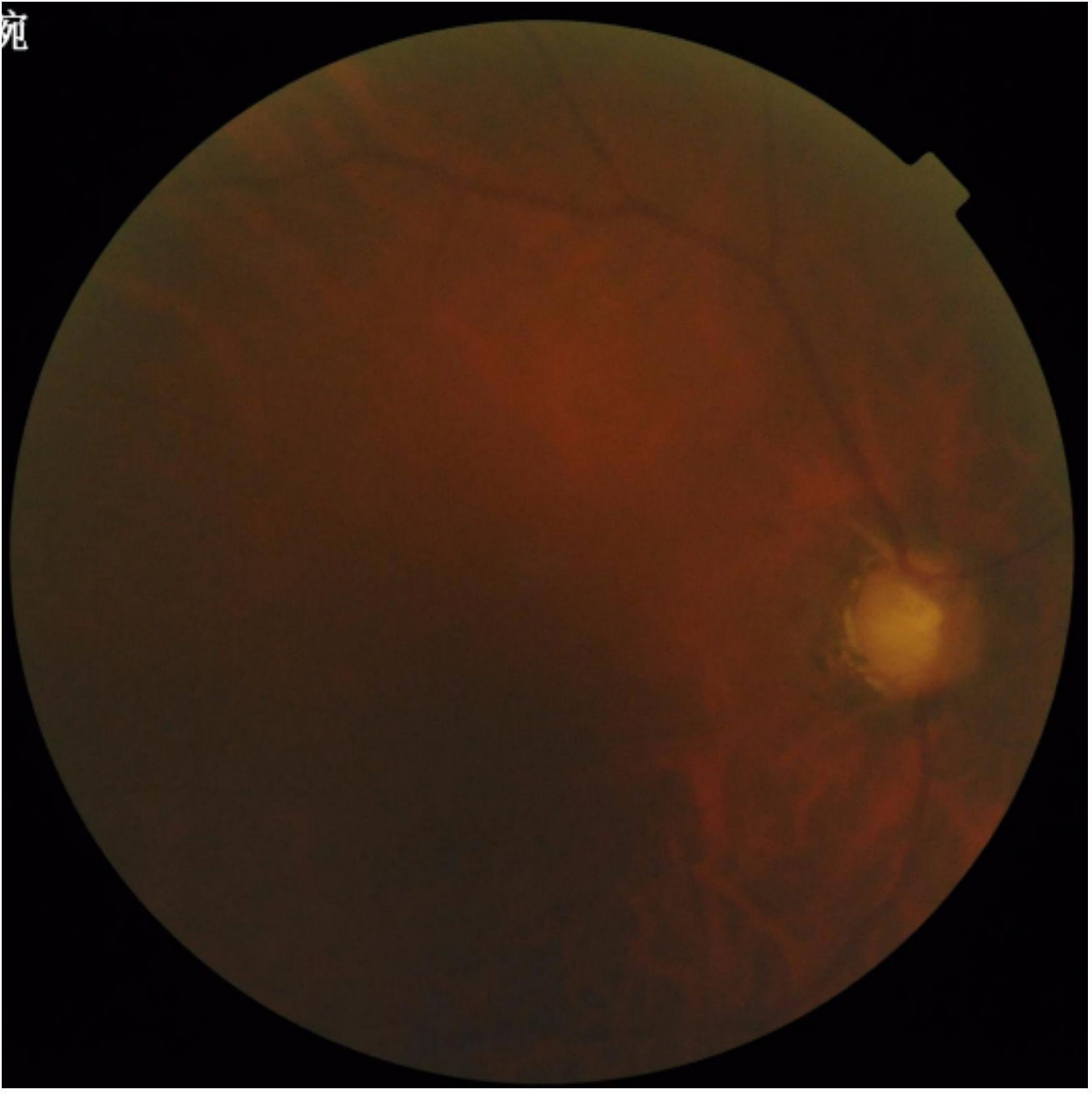} &
        \includegraphics[width=2cm]{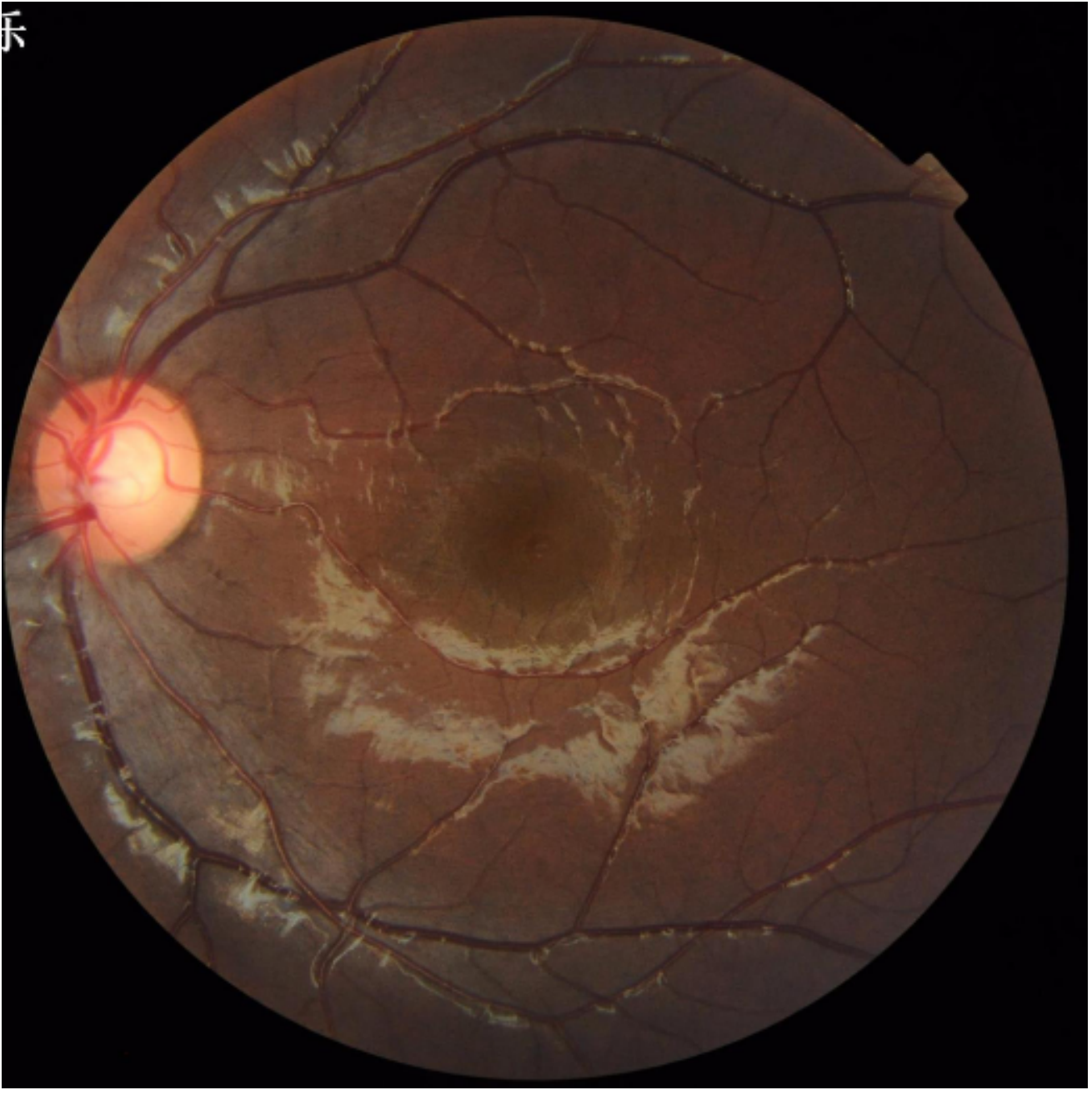} \\
        \rotatebox{90}{GT} & \includegraphics[width=2cm]{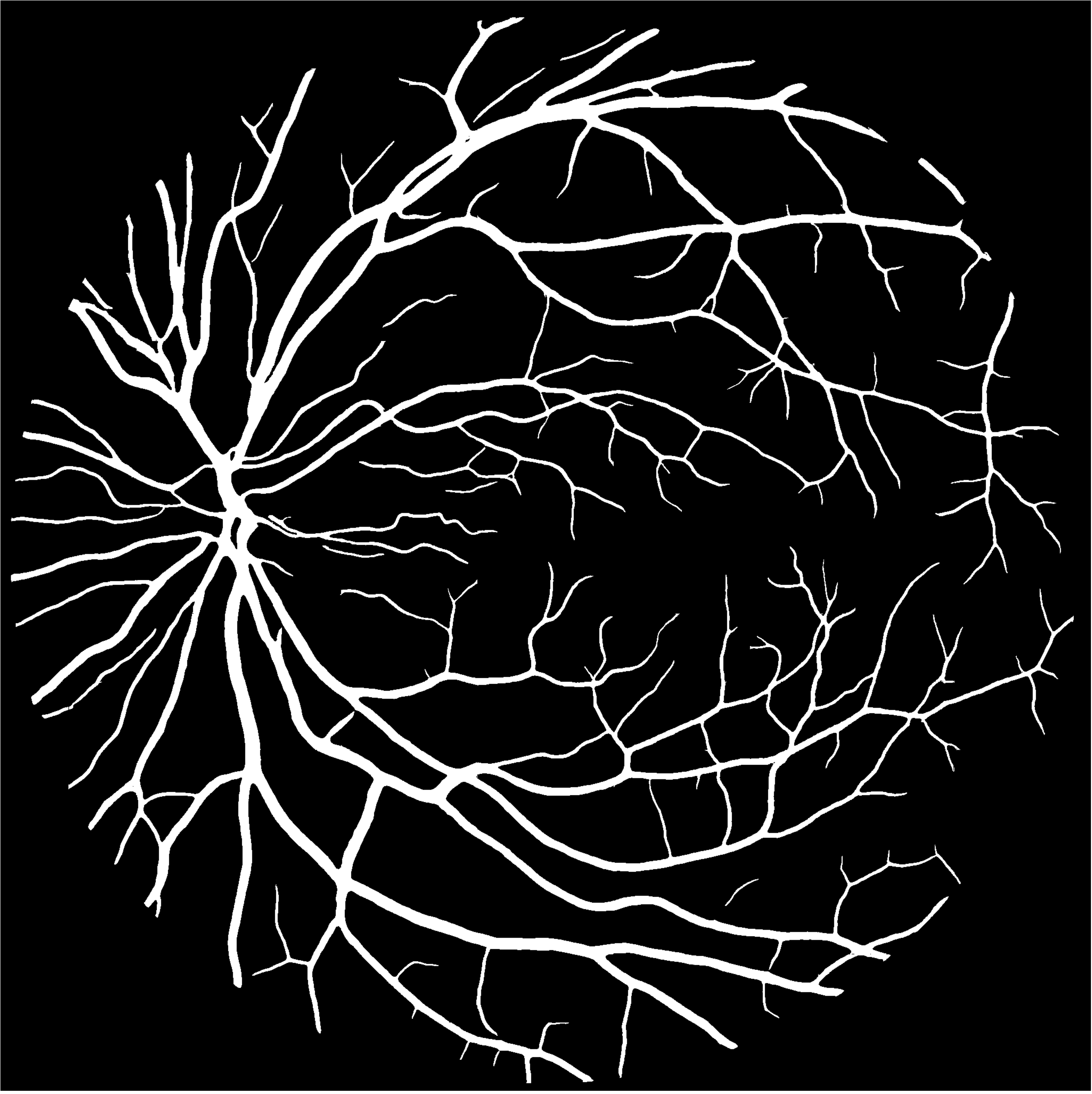} & \includegraphics[width=2cm]{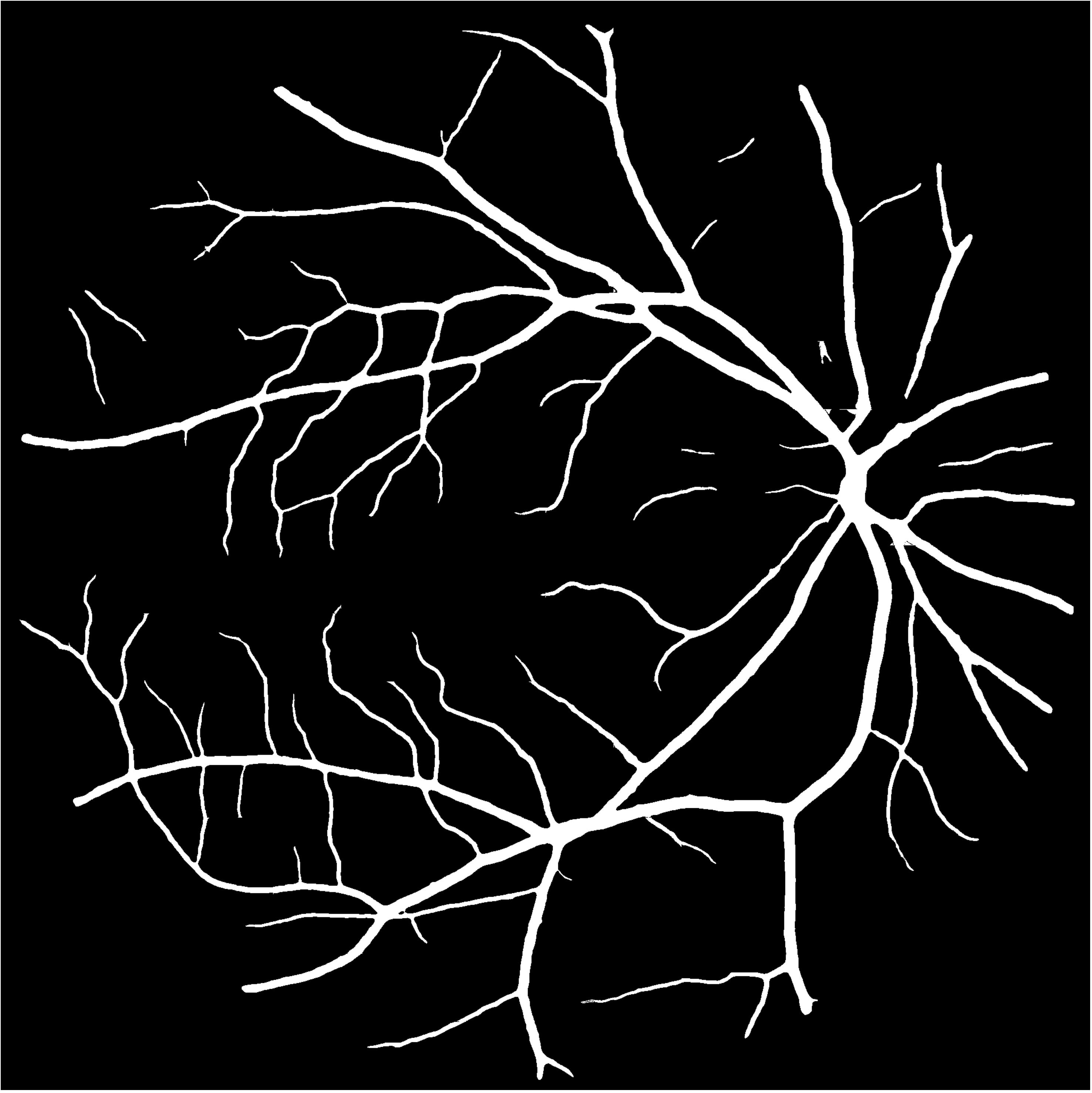} &
        \includegraphics[width=2cm]{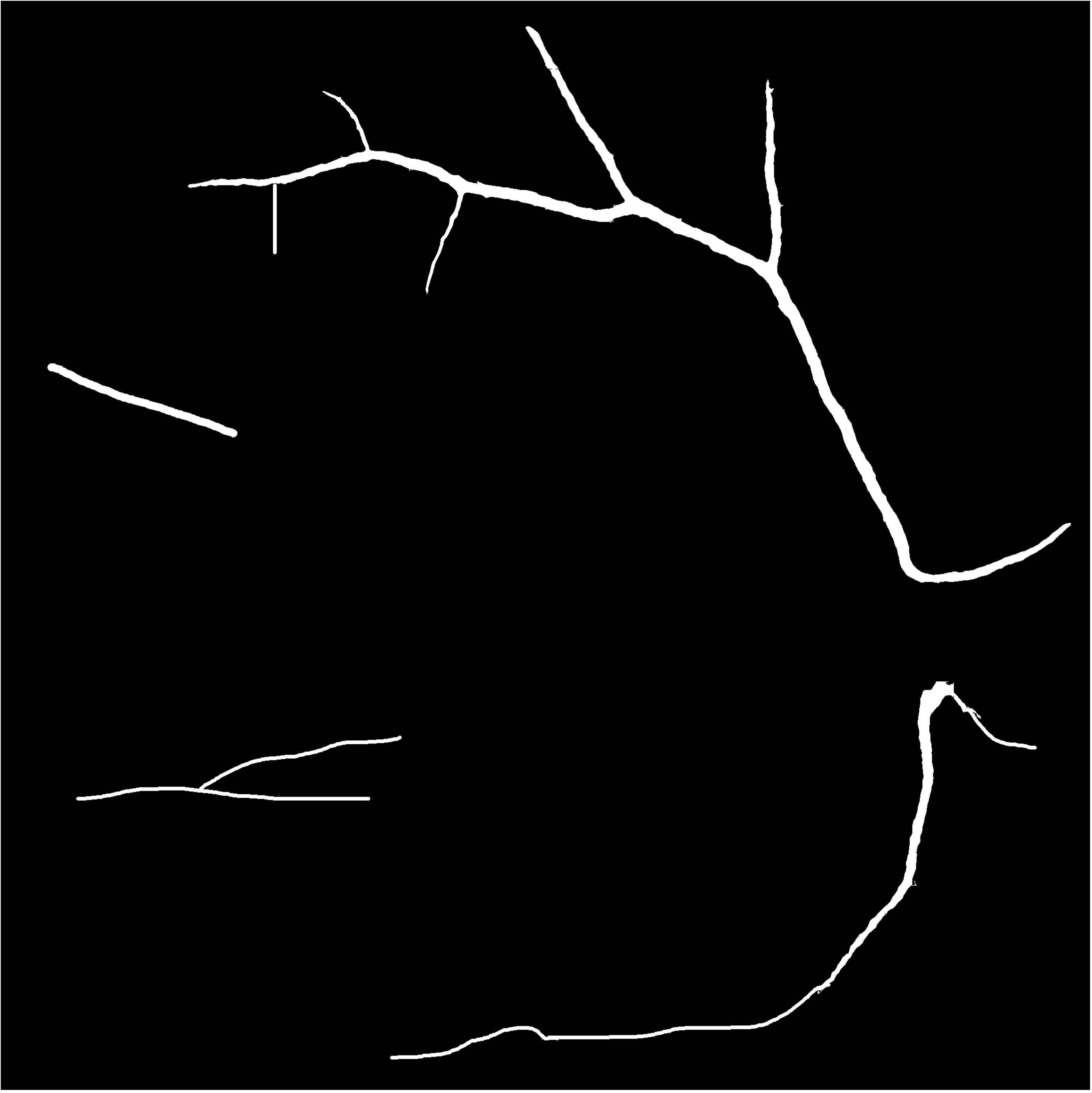} &
        \includegraphics[width=2cm]{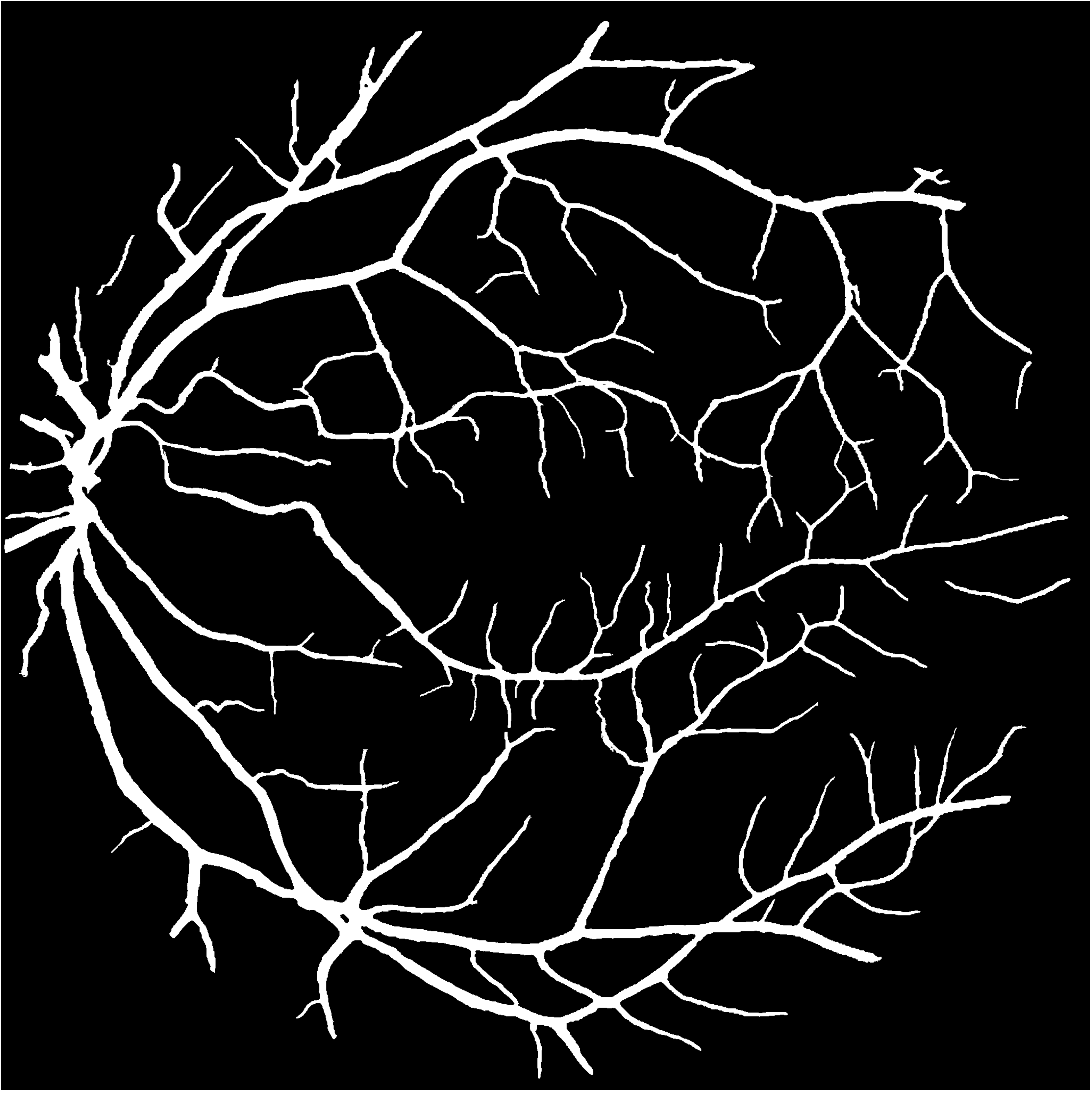} \\
        \rotatebox{90}{Pred} & \includegraphics[width=2cm]{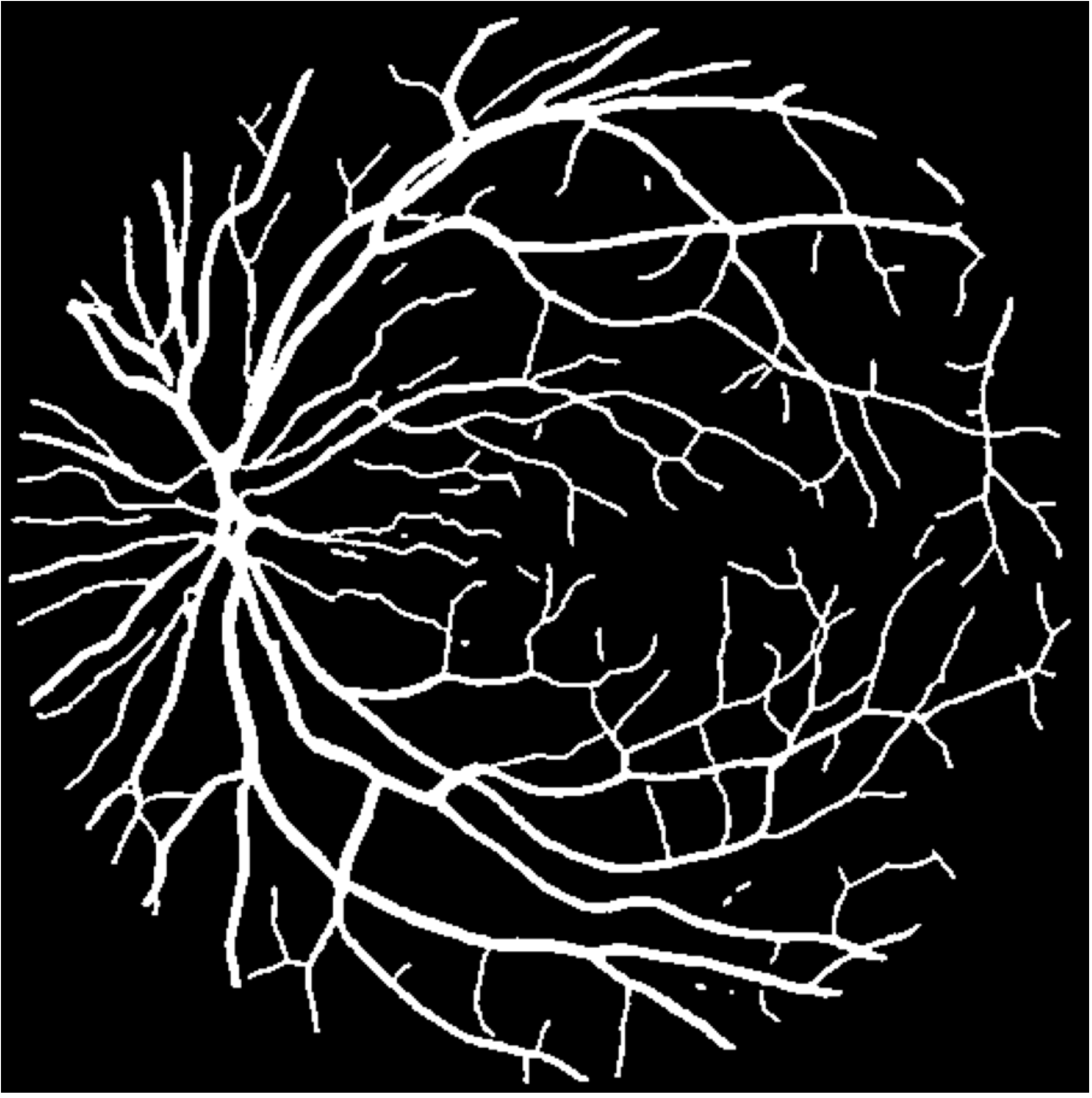} & \includegraphics[width=2cm]{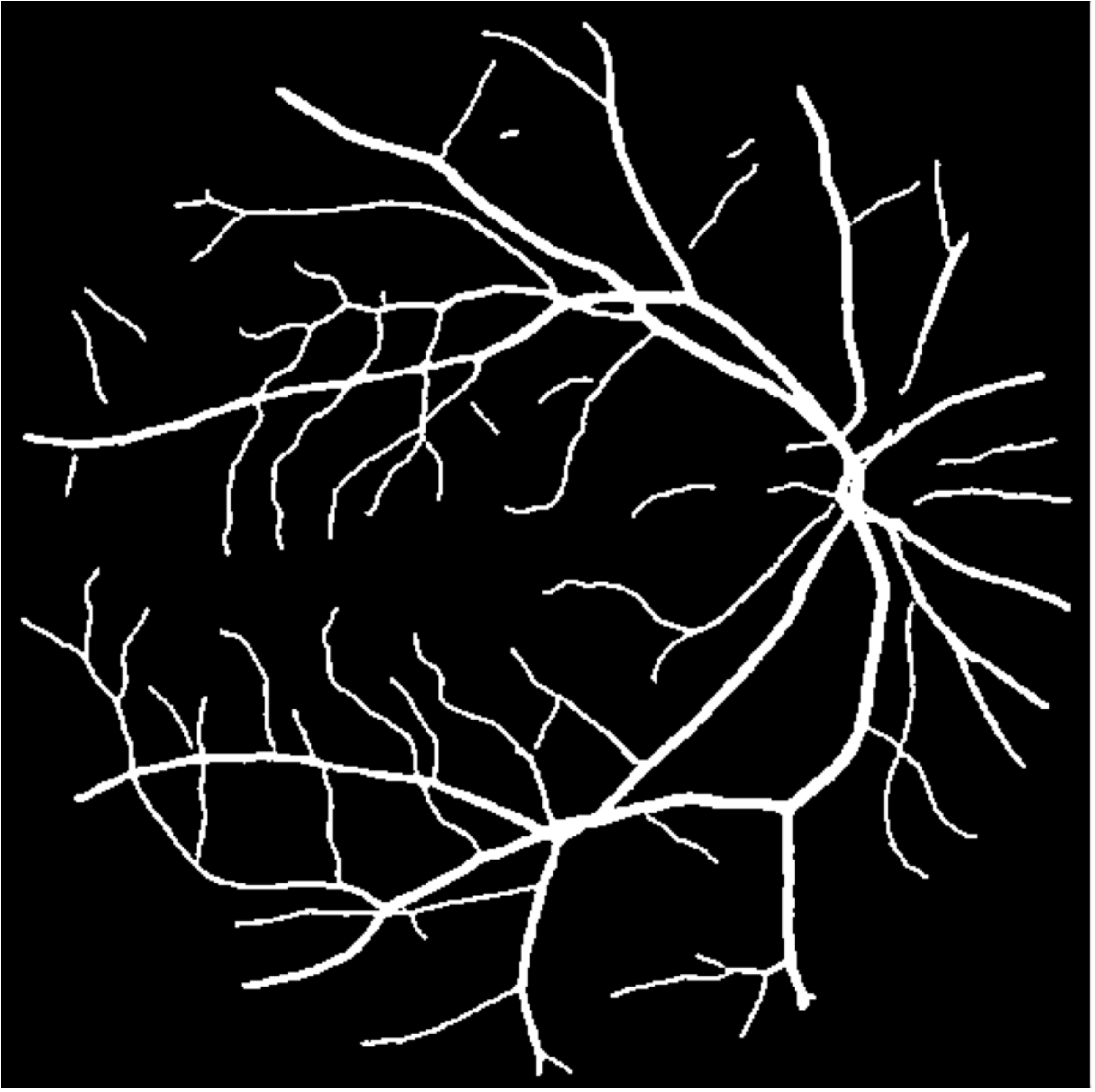} &
        \includegraphics[width=2cm]{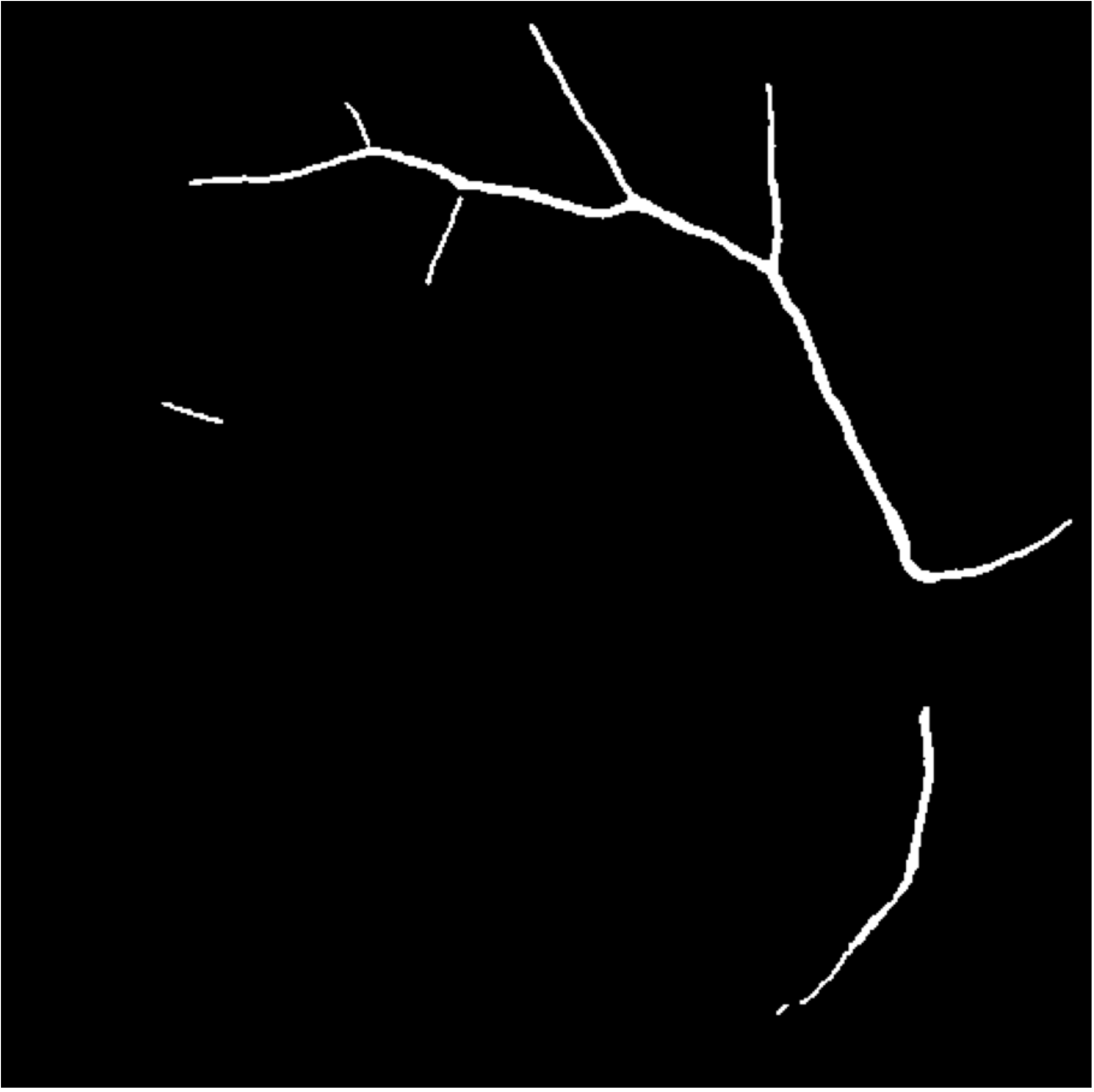} &
        \includegraphics[width=2cm]{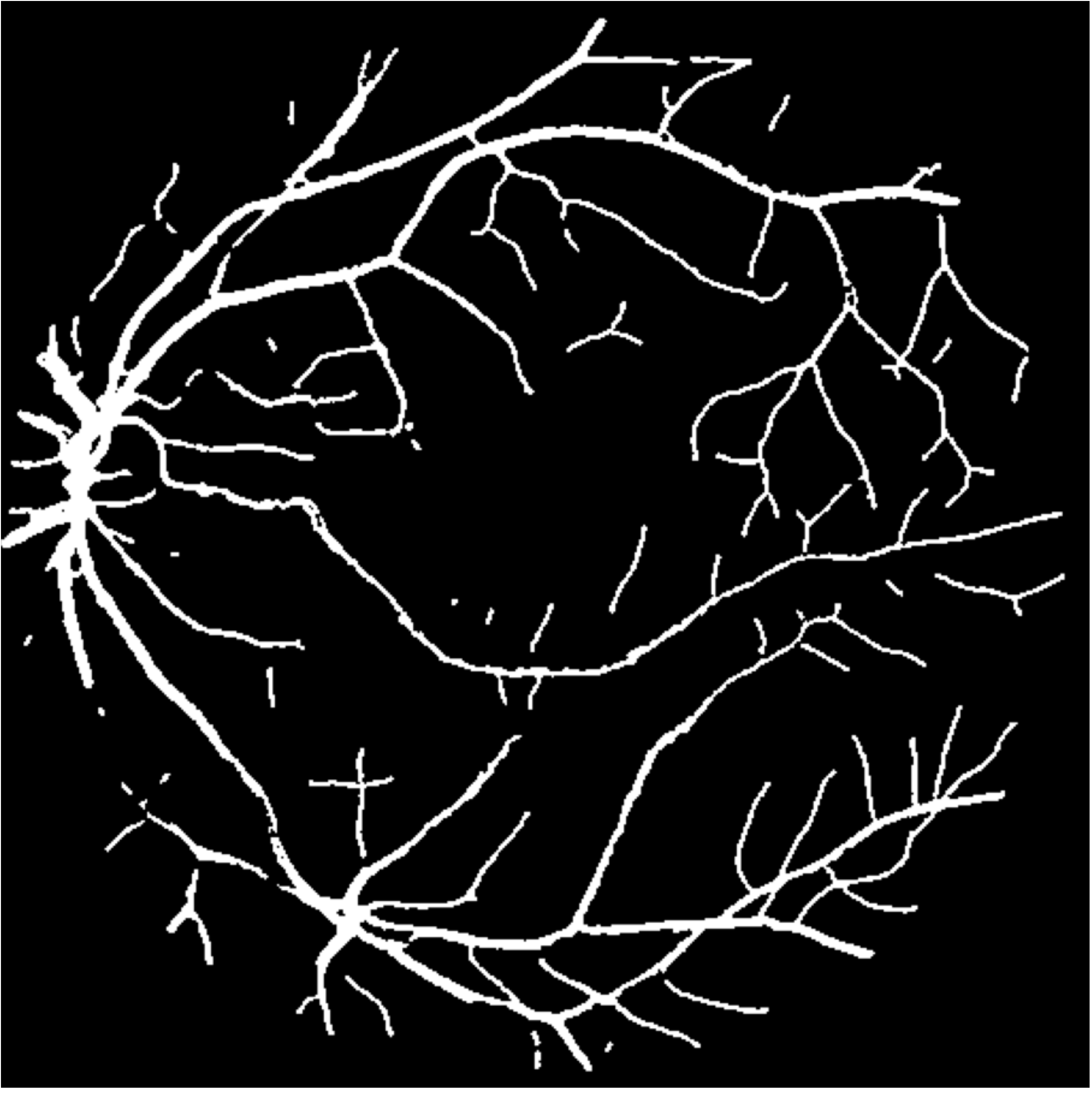} \\
    \end{tabular}
\end{table}

In our ablation studies \autoref{tabs:ablation}, we analyze the effect of the objective function and patch size on the inferences. We first note the strong effect of clDice \cite{shit2021cldice} in both pixel-wise Metrics and connectivity metrics. Given that clDice \cite{shit2021cldice} is specifically designed for tubular structure, it can be inferred that increasing the patch size causes over-segmentation and damages the center-line connectivity and therefore the segmentation results. This can be observed in \autoref{tabs:ablation}, where the 2$\times$2 patching actually worsens the results. 


\begin{table}[t]
\caption{Ablation study on different loss functions and patch sizes on FIVES \cite{jin2022fives}.} 
\label{tabs:ablation}\vspace{-10pt}
\resizebox{\textwidth}{!}{
\centering
\begin{tabular}{|l|c|ccc|cccc|}
\hline
 & & \multicolumn{3}{c|}{\segmentation{}} & \multicolumn{4}{c|}{\continuity{}} \\  \cline{3-9}
\multirow{4}{*}{\rotatebox{90}{Loss}}  & & Pre (\%) $\uparrow$ & Rec (\%) $\uparrow$ & Dice(\%) $\uparrow$ & clDice (\%) $\uparrow$ & $error_{\beta_{0}}$ $\downarrow$ & $error_{\beta_{1}}$ $\downarrow$ & $error_{\chi}$ $\downarrow$\\  \cline{3-9}

 & CE Loss            &        0.89        &  0.77  &  0.81       &      0.78       &       52.6 & 4.4            &      53.6  \\    
& CE + clDice & \textbf{0.90}  & 0.81 & \textbf{0.85} & 0.84  &    22.7 & 4.2                &     27.1  \\ 
 & clDice & \textbf{0.90} & \textbf{0.85} & \textbf{0.85} & \textbf{0.87} & \textbf{12.1} & \textbf{3.7} & \textbf{14.4} \\ %
 \hline \hline
 \multirow{2}{*}{n} & 1$\times$1      &      \textbf{0.90}          &       \textbf{0.85}  & \textbf{0.85}  &     \textbf{0.87}         &        \textbf{ 12.1}          & \textbf{ 3.7 }   & \textbf{14.4 }\\  
& 2$\times$2      &       0.71         &  0.84        & 0.76  &  0.83            &       12.83            &  5.13    & 17.37\\  \hline
\end{tabular}
}
\end{table}

\section{Conclusion}
In conclusion, preserving shape continuity is critical in medical imaging as it can impact the accuracy of medical diagnoses. Unfortunately, traditional deep learning methods often neglect this aspect, leading to inaccurate predictions. In this work, we proposed a graph-based approach that enforces shape continuity in medical segmentation by encoding it as a graph constraint. Our method significantly improved shape continuity metrics compared to traditional methods while maintaining or improving segmentation performance. By enforcing shape continuity, our approach can potentially improve the accuracy of medical diagnoses, especially in structures such as vessels, where continuity is critical.

%
%
\bibliographystyle{unsrt}
\bibliography{references}

\begin{thebibliography}{10}

\bibitem{guo2021sa}
Changlu Guo, M{\'a}rton Szemenyei, Yugen Yi, Wenle Wang, Buer Chen, and Changqi
  Fan.
\newblock Sa-unet: Spatial attention u-net for retinal vessel segmentation.
\newblock In {\em 2020 25th international conference on pattern recognition
  (ICPR)}, pages 1236--1242. IEEE, 2021.

\bibitem{kamran2021rv}
Sharif~Amit Kamran, Khondker~Fariha Hossain, Alireza Tavakkoli, Stewart~Lee
  Zuckerbrod, Kenton~M Sanders, and Salah~A Baker.
\newblock Rv-gan: Segmenting retinal vascular structure in fundus photographs
  using a novel multi-scale generative adversarial network.
\newblock In {\em International Conference on Medical Image Computing and
  Computer-Assisted Intervention}, pages 34--44. Springer, 2021.

\bibitem{zhou2021study}
Yuqian Zhou, Hanchao Yu, and Humphrey Shi.
\newblock Study group learning: Improving retinal vessel segmentation trained
  with noisy labels.
\newblock In {\em Medical Image Computing and Computer Assisted
  Intervention--MICCAI 2021: 24th International Conference, Strasbourg, France,
  September 27--October 1, 2021, Proceedings, Part I 24}, pages 57--67.
  Springer, 2021.

\bibitem{yan2018joint}
Zengqiang Yan, Xin Yang, and Kwang-Ting Cheng.
\newblock Joint segment-level and pixel-wise losses for deep learning based
  retinal vessel segmentation.
\newblock {\em IEEE Transactions on Biomedical Engineering}, 65(9):1912--1923,
  2018.

\bibitem{shit2021cldice}
Suprosanna Shit, Johannes~C Paetzold, Anjany Sekuboyina, Ivan Ezhov, Alexander
  Unger, Andrey Zhylka, Josien~PW Pluim, Ulrich Bauer, and Bjoern~H Menze.
\newblock cldice-a novel topology-preserving loss function for tubular
  structure segmentation.
\newblock In {\em Proceedings of the IEEE/CVF Conference on Computer Vision and
  Pattern Recognition}, pages 16560--16569, 2021.

\bibitem{clough2020topologicalBetti}
James~R Clough, Nicholas Byrne, Ilkay Oksuz, Veronika~A Zimmer, Julia~A
  Schnabel, and Andrew~P King.
\newblock A topological loss function for deep-learning based image
  segmentation using persistent homology.
\newblock {\em IEEE Transactions on Pattern Analysis and Machine Intelligence},
  44(12):8766--8778, 2020.

\bibitem{zhang2022topology}
Han Zhang and Lok~Ming Lui.
\newblock Topology-preserving segmentation network: A deep learning
  segmentation framework for connected component.
\newblock {\em arXiv preprint arXiv:2202.13331}, 2022.

\bibitem{zhang2018road}
Zhengxin Zhang, Qingjie Liu, and Yunhong Wang.
\newblock Road extraction by deep residual u-net.
\newblock {\em IEEE Geoscience and Remote Sensing Letters}, 15(5):749--753,
  2018.

\bibitem{alom2019recurrent}
Md~Zahangir Alom, Chris Yakopcic, Mahmudul Hasan, Tarek~M Taha, and Vijayan~K
  Asari.
\newblock Recurrent residual u-net for medical image segmentation.
\newblock {\em Journal of Medical Imaging}, 6(1):014006--014006, 2019.

\bibitem{liu2022full}
Wentao Liu, Huihua Yang, Tong Tian, Zhiwei Cao, Xipeng Pan, Weijin Xu, Yang
  Jin, and Feng Gao.
\newblock Full-resolution network and dual-threshold iteration for retinal
  vessel and coronary angiograph segmentation.
\newblock {\em IEEE Journal of Biomedical and Health Informatics},
  26(9):4623--4634, 2022.

\bibitem{zhuang2018laddernet}
Juntang Zhuang.
\newblock Laddernet: Multi-path networks based on u-net for medical image
  segmentation.
\newblock {\em arXiv preprint arXiv:1810.07810}, 2018.

\bibitem{li2020iternet}
Liangzhi Li, Manisha Verma, Yuta Nakashima, Hajime Nagahara, and Ryo Kawasaki.
\newblock Iternet: Retinal image segmentation utilizing structural redundancy
  in vessel networks.
\newblock In {\em Proceedings of the IEEE/CVF winter conference on applications
  of computer vision}, pages 3656--3665, 2020.

\bibitem{wyburd2021teds}
Madeleine~K Wyburd, Nicola~K Dinsdale, Ana~IL Namburete, and Mark Jenkinson.
\newblock Teds-net: enforcing diffeomorphisms in spatial transformers to
  guarantee topology preservation in segmentations.
\newblock In {\em Medical Image Computing and Computer Assisted
  Intervention--MICCAI 2021: 24th International Conference, Strasbourg, France,
  September 27--October 1, 2021, Proceedings, Part I 24}, pages 250--260.
  Springer, 2021.

\bibitem{shin2019deep}
Seung~Yeon Shin, Soochahn Lee, Il~Dong Yun, and Kyoung~Mu Lee.
\newblock Deep vessel segmentation by learning graphical connectivity.
\newblock {\em Medical image analysis}, 58:101556, 2019.

\bibitem{velivckovic2017graph}
Petar Veli{\v{c}}kovi{\'c}, Guillem Cucurull, Arantxa Casanova, Adriana Romero,
  Pietro Lio, and Yoshua Bengio.
\newblock Graph attention networks.
\newblock {\em arXiv preprint arXiv:1710.10903}, 2017.

\bibitem{9562259}
Ruikun Li, Yi-Jie Huang, Huai Chen, Xiaoqing Liu, Yizhou Yu, Dahong Qian, and
  Lisheng Wang.
\newblock 3d graph-connectivity constrained network for hepatic vessel
  segmentation.
\newblock {\em IEEE Journal of Biomedical and Health Informatics},
  26(3):1251--1262, 2022.

\bibitem{yu2022vessel}
Hao Yu, Jie Zhao, and Li~Zhang.
\newblock Vessel segmentation via link prediction of graph neural networks.
\newblock In {\em Multiscale Multimodal Medical Imaging: Third International
  Workshop, MMMI 2022, Held in Conjunction with MICCAI 2022, Singapore,
  September 22, 2022, Proceedings}, pages 34--43. Springer, 2022.

\bibitem{ronneberger2015u}
Olaf Ronneberger, Philipp Fischer, and Thomas Brox.
\newblock U-net: Convolutional networks for biomedical image segmentation.
\newblock In {\em Medical Image Computing and Computer-Assisted
  Intervention--MICCAI 2015: 18th International Conference, Munich, Germany,
  October 5-9, 2015, Proceedings, Part III 18}, pages 234--241. Springer, 2015.

\bibitem{simonyan2014very}
Karen Simonyan and Andrew Zisserman.
\newblock Very deep convolutional networks for large-scale image recognition.
\newblock {\em arXiv preprint arXiv:1409.1556}, 2014.

\bibitem{maninis2016deep}
Kevis-Kokitsi Maninis, Jordi Pont-Tuset, Pablo Arbel{\'a}ez, and Luc Van~Gool.
\newblock Deep retinal image understanding.
\newblock In {\em Medical Image Computing and Computer-Assisted
  Intervention--MICCAI 2016: 19th International Conference, Athens, Greece,
  October 17-21, 2016, Proceedings, Part II 19}, pages 140--148. Springer,
  2016.

\bibitem{scarselli2008graph}
Franco Scarselli, Marco Gori, Ah~Chung Tsoi, Markus Hagenbuchner, and Gabriele
  Monfardini.
\newblock The graph neural network model.
\newblock {\em IEEE transactions on neural networks}, 20(1):61--80, 2008.

\bibitem{kipf2016semi}
Thomas~N Kipf and Max Welling.
\newblock Semi-supervised classification with graph convolutional networks.
\newblock {\em arXiv preprint arXiv:1609.02907}, 2016.

\bibitem{fraz2012ensemble}
Muhammad~Moazam Fraz, Paolo Remagnino, Andreas Hoppe, Bunyarit Uyyanonvara,
  Alicja~R Rudnicka, Christopher~G Owen, and Sarah~A Barman.
\newblock An ensemble classification-based approach applied to retinal blood
  vessel segmentation.
\newblock {\em IEEE Transactions on Biomedical Engineering}, 59(9):2538--2548,
  2012.

\bibitem{jin2022fives}
Kai Jin, Xingru Huang, Jingxing Zhou, Yunxiang Li, Yan Yan, Yibao Sun, Qianni
  Zhang, Yaqi Wang, and Juan Ye.
\newblock Fives: A fundus image dataset for artificial intelligence based
  vessel segmentation.
\newblock {\em Scientific Data}, 9(1):475, 2022.

\bibitem{kerfoot2019left}
Eric Kerfoot, James Clough, Ilkay Oksuz, Jack Lee, Andrew~P King, and Julia~A
  Schnabel.
\newblock Left-ventricle quantification using residual u-net.
\newblock In {\em Statistical Atlases and Computational Models of the Heart.
  Atrial Segmentation and LV Quantification Challenges: 9th International
  Workshop, STACOM 2018, Held in Conjunction with MICCAI 2018, Granada, Spain,
  September 16, 2018, Revised Selected Papers 9}, pages 371--380. Springer,
  2019.

\bibitem{beltramo2021euler}
Gabriele Beltramo, Rayna Andreeva, Ylenia Giarratano, Miguel~O Bernabeu, Rik
  Sarkar, and Primoz Skraba.
\newblock Euler characteristic surfaces.
\newblock {\em arXiv preprint arXiv:2102.08260}, 2021.

\end{thebibliography}
\end{document}